\documentclass[conference]{IEEEtran}
\IEEEoverridecommandlockouts

\usepackage{flushend}
\usepackage{graphics}
\usepackage{algorithmic}
\usepackage{algorithm}
\usepackage{caption}
\usepackage{tabularray}
\usepackage{array}
\usepackage{amsmath}
\usepackage{amssymb}
\usepackage{cancel}
\usepackage{esint}
\usepackage{mathdots}
\usepackage{mathtools}
\usepackage{mhchem}
\usepackage{stackrel}
\usepackage{stmaryrd}

\usepackage[pdfpagelayout=OneColumn, pdfnewwindow=true, pdfstartview=XYZ, plainpages=false]{hyperref}
\hypersetup{
    pdftitle={Message Passing Based Two-Timescale Bayesian Learning for Joint Channel and Memory Hardware Impairments Tracking},
    pdfauthor={Wei Xu; An Liu},
    pdfborder={0 0 1},
    bookmarksnumbered=true,
    bookmarksopen=true,
    bookmarksopenlevel=1,
    colorlinks=false,
}

\title{Message Passing Based Two-Timescale Bayesian Learning for Joint Channel and Memory Hardware Impairments Tracking}

\author{%
{\normalsize Wei Xu, }{\normalsize\textit{Graduate Student Member, IEEE,}}%
{\normalsize{} and An Liu, }{\normalsize\textit{Senior Member, IEEE}}%
\vspace{-0.2in}%
{\normalsize\thanks{This work was supported in part by the National Key Research and Development
Program of China under Grant 2025ZD1301900; in part by National Key Laboratory of Millimeter-Wave and
Terahertz Remote Sensing and the Zhejiang Provincial Key Laboratory of Information Processing, Communication
and Networking (IPCAN), Hangzhou, China; in part by the Zhejiang Provincial Key Laboratory of Multi-Modal Communication Networks and Intelligent Information Processing, Hangzhou, China. (Corresponding authors: An Liu.)\par Wei Xu and An Liu are with the College of Information
Science and Electronic Engineering, Zhejiang University, Hangzhou
310027, China (e-mails: 12231077@zju.edu.cn and anliu@zju.edu.cn).}}%
}

\begin{document}

\maketitle

\begin{abstract}
Hardware impairments in massive multiple-input multiple-output (MIMO) receivers introduce inter-symbol memory and inter-element coupling, severely degrading channel estimation. This paper employs a residual recurrent gated unit (RGRU) to model the intra-slot memory of the hardware impairments and proposes a message-passing-based two-timescale Bayesian deep learning (MP-TTBDL) framework for joint channel and impairment tracking. Owing to small-scale fading, the wireless channel varies rapidly across slots, whereas hardware impairments drift slowly due to hardware aging and environmental variations. To capture these distinct physical timescales, a fast-varying Markov prior and a slow-varying Gaussian Markov prior are assigned to the sparse channel and the network parameters, respectively. Based on a multi-slot factor graph formulation, a message-passing algorithm is developed. Specifically, the inter-slot messages admit closed-form updates, while the intra-slot factor graph, due to its complex recurrent structure, is partitioned into a channel tracking module and an impairments calibration module. The channel tracking module performs sparse channel estimation via turbo orthogonal approximate message passing (Turbo-OAMP), and the impairments calibration module updates the impairment parameters via a specially designed deep approximate message passing (DAMP) procedure, with the two modules iteratively exchanging extrinsic information through expectation propagation (EP) until convergence. Simulation results show that the proposed framework robustly achieves lower channel estimation error than conventional compensators followed by channel estimation across different online impairment scenarios and signal-to-noise ratio (SNR) conditions.
\end{abstract}


\section{Introduction}
Massive MIMO has become a cornerstone of modern wireless communications, with base stations (BSs) deploying large antenna arrays to exploit spatial diversity and multiplexing gains.
To realize these gains, accurate channel estimation and beamforming are indispensable.
However, as the number of antennas grows, RF front-end hardware impairments including LNA nonlinearity~\cite{moghaddamStatistical2022,wangPASurvey2023}, crosstalk between RF chains~\cite{bassamCrossover2009,bassamExperimental2008}, and IQ imbalance~\cite{tarighatCompensation2005,alacaReceiverIQ2022} become increasingly severe, introducing unknown phase and amplitude errors, distorting pilot signals, and severely degrading channel estimation performance.
These impairments have gradually become a critical bottleneck in system design.

To mitigate these impairments, conventional methods typically train an offline compensator to learn the inverse mapping of the hardware impairments, and then apply the learned model with fixed parameters during online reception~\cite{zhaoLowComplexity2023,liNearField2026,qiAnalysis2011,wuDeep2023}.
Since practical RF front-end hardware exhibits both nonlinearity and memory effects across symbols~\cite{aminBehavioral2014,obrienEstimation2006}, the Volterra series has been widely used to characterize these effects and their inverses~\cite{morganGeneralized2006,kongWDM2022,goffioulVolterra2002}.
However, the number of Volterra parameters grows combinatorially with the nonlinearity order, memory depth, and the number of RF chains, making it prohibitive for large-scale systems.
As a practical alternative, simplified models such as the generalized memory polynomial (GMP)~\cite{morganGeneralized2006} are often adopted for compensation.
Yet, the exact inverse mapping may not be accurately representable by polynomials, and GMP-based compensators can therefore suffer from significant model mismatch.
Recently, recurrent neural networks such as long short-term memory (LSTM) and GRU~\cite{fawzyIterative2023,liHighRelativeBandwidth2024,jarautComposite2018} have been employed to compensate for the combined impairments, achieving superior performance over GMP in many digital predistortion tasks thanks to their universal approximation capability.

Existing compensation approaches suffer from two fundamental issues.
First, in the presence of inter-symbol memory and inter-chain crosstalk, each observation depends on multiple past transmitted symbols, and learning the inverse mapping that recovers the original signals from a mixed and overlapping set of contributions becomes an ill-posed problem, leading to poor compensation and limited generalization.
In contrast, the forward hardware-impairment model possesses a natural causal and recursive structure; it describes how hardware distorts original signals symbol by symbol, a structure that is far easier to approximate with a recurrent network.
With an accurate forward model, signal recovery can then be cast as a Bayesian inference problem together with priors, yielding more reliable results.
Second, conventional methods typically train the compensator offline, and the resulting static model may fail to track the actual hardware behavior during online operation due to temperature variations~\cite{taghikhaniTemperatureDependent2022}, component aging, and other environmental factors~\cite{zhengOnline2021}. Online adaptation of the impairment-aware receiver can mitigate such drift by learning from live data~\cite{kwonMIMO2024}, but most existing pipelines perform point estimation without an adaptive mechanism to regulate the update step size, and typically require buffering additional calibration samples and executing a separate retraining stage after coarse detection, which introduces substantial memory and computational overhead beyond the main receiver processing.

Channel estimation error is critical to wireless communication system performance.
In recent years, Bayesian inference has been widely applied to various channel estimation scenarios and has achieved superior estimation performance.
For example, subspace-constrained variational Bayesian inference (SC-VBI) has been proposed for structured compressive sensing with a dynamic grid by replacing high-dimensional matrix inversions with low-dimensional subspace-constrained updates~\cite{zhouSCVBI2025}, an alternating maximum a posteriori (MAP) framework jointly detects visibility regions and estimates channels in near-field spatially non-stationary extra-large MIMO (XL-MIMO) systems~\cite{xuXLMMIMO2024}, and a Turbo-CS-based sparse-Markov scheme enables robust multi-user uplink channel tracking in 5G New Radio (NR) under hopping sounding reference signal (SRS) patterns and system imperfections~\cite{wanNR2024}.
However, few existing works consider channel estimation under non-ideal transceivers with hardware impairments.

Message-passing-based Bayesian inference has also been widely applied to symbol detection, where embedding neural impairment models in a factor graph yields significant performance gains when the channel is assumed known~\cite{gaoSignal2024,gaoBayesianISAC2026}, validating the benefit of principled inference over a compensator.
Nevertheless, such approaches typically neglect the memory effects of hardware impairments, do not account for time-varying impairments during online operation, and do not consider multipath channel estimation in practical wireless communication scenarios, leaving the additional gains from jointly inferring the sparse multipath channel and impairment parameters in a unified factor graph unexploited.
Therefore, it is imperative to establish a joint Bayesian estimation framework that accounts for both the wireless channel and hardware impairments, whereas directly applying conventional compensation and then estimating the channel from the compensated signal inevitably leads to significant degradation in channel estimation performance.

Building upon our prior work that considered the joint tracking problem under memoryless impairments~\cite{xuMPTTBDL2025}, this paper further investigates the uplink channel estimation problem for a massive MIMO receiver with memory hardware impairments.
We use an RGRU to model the forward impairment mapping, and integrate sparse channel estimation with RGRU-based impairment learning in a unified factor graph.
Then we propose the MP-TTBDL framework that jointly tracks the fast-varying channel and the slow-varying impairments.

The main contributions of this paper are summarized as follows:
\begin{itemize}
    \item \textbf{A Unified Bayesian Framework for Joint Channel and Memory Impairment Estimation:}
    We employ a residual RGRU with probit activation functions to model the memory hardware impairments at the receiver. By embedding this RGRU as the observation likelihood in a unified factor graph together with the sparse channel model, we formulate the problem as a closed-loop joint Bayesian inference problem in which the channel and impairment parameters are estimated jointly from distorted pilots. By exploiting the natural causal and recursive structure of the forward impairment process, this framework avoids the ill-posed inverse mapping learned by conventional compensators and removes the need for a separately trained inverse compensator.

    \item \textbf{Two-Timescale Markov Transition Models for Joint Tracking:}
    The wireless channel varies rapidly across slots due to small-scale fading, whereas hardware impairments drift slowly owing to temperature variations and component aging. To reflect these distinct physical timescales, we design the transition probabilities of the sparse angular-domain channel as a fast-varying Markov chain, and those of the RGRU impairment parameters as a slow-varying Gaussian Markov chain. This two-timescale Markov transition model design aligns the statistical model with the practical transmission system and allows the receiver to continuously track both channel variations and hardware aging in a principled manner.

    \item \textbf{The Proposed MP-TTBDL Algorithm:}
    Based on the proposed Bayesian tracking framework, we develop the MP-TTBDL algorithm that combines Turbo-OAMP for sparse channel tracking with a novel DAMP scheme tailored to the RGRU architecture. Given the complex factor graph structure of the RGRU, we design a dedicated DAMP procedure for message updating that specifies a coordinated message update schedule and closed-form approximation methods. Within each slot, the two modules iteratively exchange extrinsic information via EP, and the updated messages are propagated across slots under the two-timescale Markov priors, enabling continuous online joint tracking.

    \item \textbf{Simulation Validation:}
    Extensive simulations are conducted to validate the proposed contributions. On synthetically generated impairment data, the dedicated DAMP converges faster and achieves higher fitting accuracy than conventional optimizers for learning the RGRU parameters. When the online impairments match the offline pretraining configuration, the proposed joint Bayesian framework yields significantly lower channel estimation normalized mean squared error (NMSE) than compensation-based baselines across a wide signal-to-noise ratio (SNR) range, validating the advantage of the joint Bayesian framework. Under slowly time-varying impairments, the MP-TTBDL algorithm maintains robust channel tracking performance, confirming the effectiveness of the two-timescale Markov prior and the MP-TTBDL algorithm.
\end{itemize}

The remainder of this paper is organized as follows.
Section~II presents the system model, the RGRU-based impairment surrogate, and the two-timescale Markov priors.
Section~III details the MP-TTBDL framework, including the factor graph design and the inference algorithm.
Section~IV provides simulation results, and Section~V concludes the paper.

Notations: $\boldsymbol{0}_{N}$ refers to an all-zero column vector of dimension $N$.
$\boldsymbol{I}_{N}$ refers to an $N\times N$ identity matrix.
$\overline{\left(\cdot\right)}$, $\left(\cdot\right)^{\mathrm{T}}$, $\left(\cdot\right)^{\mathrm{H}}$, and $\left(\cdot\right)^{\mathrm{-1}}$ denote conjugate, transpose, conjugate transpose, and inverse, respectively.
$\Re(\cdot)$ denotes the real part of a complex argument.
$\mathbb{R}$ and $\mathbb{C}$ refer to the sets of real and complex numbers.
$|a|$ and $\|\boldsymbol{a}\|_{2}$ denote the amplitude of scalar $a$ and the 2-norm of vector $\boldsymbol{a}$, respectively.
$a_{n}$ and $A_{mn}$ refer to the $n$-th entry of vector $\boldsymbol{a}$ and the $(m,n)$-th entry of matrix $\boldsymbol{A}$, respectively.
$\left[\cdot\right]_{i,j}$ denotes the $(i,j)$-th entry of a matrix.
The Hadamard (element-wise) product is denoted by $\odot$, and the Kronecker product is denoted by $\otimes$.
$\mathcal{O}(\cdot)$ denotes the order of complexity.
$o(\cdot)$ denotes higher-order infinitesimals.
$\mathcal{R}(\cdot)=\begin{bmatrix} \Re(\cdot) \\ \Im(\cdot) \end{bmatrix}$ facilitates computation in the real domain.

\section{System Model}
\subsection{Ideal Uplink Pilot Signal Model}
In a 5G wideband communication system, a massive MIMO BS equipped with $N$ antennas serves multiple single-antenna users. Without loss of generality, we focus on the uplink channel estimation problem for a single user under flat fading. The proposed framework admits straightforward extensions to frequency-selective channels and multi-user scenarios. We denote by $P$ the number of pilot symbols per slot and by $p\in\{1,\ldots,P\}$ the pilot symbol index. Then, the received signal $\boldsymbol{y}_{p}(t)\in\mathbb{C}^{N}$ at symbol $p$ of slot $t$ is expressed as:
\begin{equation}
\boldsymbol{y}_{p}\left(t\right)=\boldsymbol{h}\left(t\right)r_{p}\left(t\right)+\boldsymbol{n}_{p}\left(t\right),\quad p=1,\ldots,P,\label{eq:obs}
\end{equation}
where $r_{p}(t)\in\mathbb{C}$ is the uplink pilot symbol at symbol $p$, $\boldsymbol{h}(t)\in\mathbb{C}^{N}$ is the channel vector assumed constant within slot $t$, and $\boldsymbol{n}_{p}(t)\in\mathbb{C}^{N}$ is the additive white Gaussian noise (AWGN).

Due to sparse multipath propagation, we express the channel $\boldsymbol{h}(t)$ as a linear combination of off-grid angular-domain basis vectors~\cite{lianExploiting2019} to exploit angular-domain sparsity for improved channel estimation accuracy:
\[
\boldsymbol{h}=\boldsymbol{A}\left(\boldsymbol{\vartheta}\right)\boldsymbol{x},
\]
where $\boldsymbol{\vartheta}\in\mathbb{R}^{N}$ is a dynamic grid with $N$ grid points in angular domain, $\boldsymbol{x}\in\mathbb{C}^{N}$ is the sparse angular-domain channel vector, and columns in $\boldsymbol{A}\left(\boldsymbol{\vartheta}\right)$ are steering vectors
\begin{align*}
\boldsymbol{A}\left(\boldsymbol{\vartheta}\right)&=\left[\boldsymbol{a}\left(\vartheta_{1}\right),\ldots,\boldsymbol{a}\left(\vartheta_{N}\right)\right], \\
\boldsymbol{a}\left(\vartheta\right)&\triangleq\left[1,e^{j\pi\sin\left(\vartheta\right)},\ldots,e^{j\left(N-1\right)\pi\sin\left(\vartheta\right)}\right]^{T}.
\end{align*}
For clarity, we adopt the half-wavelength spaced uniform linear array (ULA) steering-vector model above. Extension to other array geometries, such as uniform planar arrays (UPA), is straightforward by replacing the steering-vector dictionary $\boldsymbol{A}(\boldsymbol{\vartheta})$ with the corresponding array response, while the overall Bayesian inference and message-passing framework remains unchanged.
The dynamic angle grid $\boldsymbol{\vartheta}$ is often initialized using a uniform grid and then updated based on gradient method.

\subsection{Uplink Signal Model with Memory Hardware Impairments}
\label{subsec:Practical-Transmission-Model}

\begin{figure}[!t]
\centering
\includegraphics[width=\linewidth]{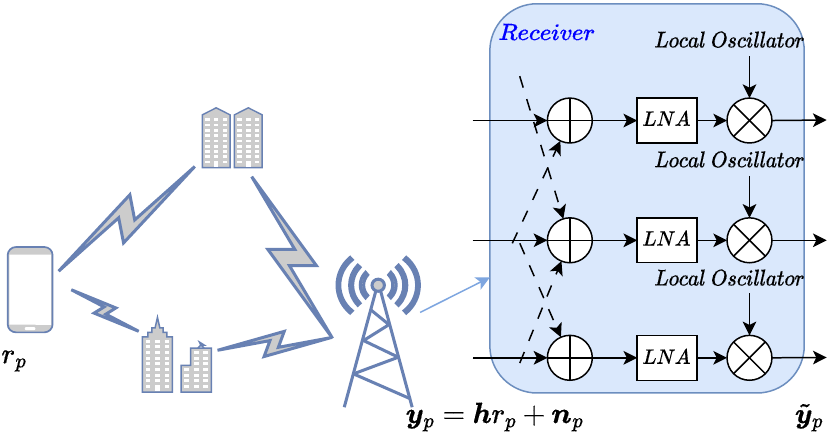}
\caption{Block diagram of a practical massive MIMO uplink receiver with hardware impairments; RF-chain crosstalk, LNA nonlinearity, and IQ imbalance are shown as representative examples.}
\label{fig:system-model}
\end{figure}

In a practical massive MIMO receiver, the signal captured by the antenna array is inevitably distorted by the non-ideal RF front-end. As illustrated in Fig.~\ref{fig:system-model}, the hardware impairments can take various forms. For instance: (i) electromagnetic coupling among closely spaced RF chains introduces linear and nonlinear crosstalk; (ii) the LNA exhibits nonlinear amplification with memory, where the output depends on current and several past inputs; and (iii) IQ imbalance arising from mismatched local oscillator branches results in phase and amplitude distortion of the complex baseband signal~\cite{qiAnalysis2011,zhaoLowComplexity2023,uthayakumarScalable2023}. Together, these effects impose a nonlinear mapping with inter-symbol memory and inter-element coupling on the received pilot, making hardware distortion a severe bottleneck for channel estimation as the antenna count grows.

In prior digital predistortion works, such memory input-output behavior is typically modeled by memory polynomials~\cite{morganGeneralized2006} or recurrent neural networks~\cite{fawzyIterative2023,liHighRelativeBandwidth2024,jarautComposite2018}, where GRU-type networks in particular achieve robustly low fitting error. Besides, neural networks are more general in modeling nonlinear mappings and can be used to characterize more complicated hardware impairments in practice. In such tasks, a single-layer RGRU already provides sufficient modeling accuracy, as demonstrated in~\cite{wuOpenDPD2024}; deeper RGRUs, due to the large number of extra parameters, tend to overfit the training data and suffer performance degradation during online deployment. Therefore, in this paper, we only consider a single-layer RGRU. For each slot $t$, let $\boldsymbol{\pi}_{p}(t)\in\mathbb{R}^{N_{\pi}}$ denote the hidden state of the RGRU after processing the $p$-th pilot symbol. Given $\boldsymbol{y}_{p}(t)$ and $\boldsymbol{\pi}_{p}(t)$, the RGRU outputs an estimate of $\mathcal{R}(\tilde{\boldsymbol{y}}_{p}(t))$ with AWGN:
\begin{equation}
\mathcal{R}\left(\tilde{\boldsymbol{y}}_{p}\left(t\right)\right)=\mathcal{R}\left(\boldsymbol{y}_{p}\left(t\right)\right)+\boldsymbol{W}_{\tilde{y}}\left(t\right)\boldsymbol{\pi}_{p}\left(t\right)+\boldsymbol{b}_{\tilde{y}}\left(t\right)+\tilde{\boldsymbol{n}}_{p},\label{eq:resnet}
\end{equation}
where we employ $\mathcal{R}(\cdot)$ to facilitate computing in the real domain, and denote the weights/bias in the output layer as $\boldsymbol{W}_{\tilde{y}}(t)$/$\boldsymbol{b}_{\tilde{y}}(t)$, respectively. $\tilde{\boldsymbol{n}}_{p}\sim\prod_{n=1}^{2N}\mathcal{N}(n_{n};0,\tilde{\sigma}^{2})$ represents the fitting error. The hidden state $\boldsymbol{\pi}_{p}(t)$ encodes the symbol-level impairment memory accumulated along the pilot chain within slot $t$. It is recurrently updated by a single-layer GRU: $\boldsymbol{z}_{p}(t)$ and $\boldsymbol{c}_{p}(t)$ are the update and reset gates, $\check{\boldsymbol{\pi}}_{p}(t)$ filters the past memory, and $\boldsymbol{\tilde{\pi}}_{p}(t)$ extracts a new memory component from the current real-domain pilot $\mathcal{R}(\boldsymbol{y}_{p}(t))$:
\begin{flalign}
& \boldsymbol{z}_{p}\left(t\right) =\boldsymbol{W}_{z}\left(t\right)[\boldsymbol{\pi}_{p-1}\left(t\right);\mathcal{R}\left(\boldsymbol{y}_{p}\left(t\right)\right)]+\boldsymbol{b}_{z}\left(t\right), && \label{eq:zt}\\
& \boldsymbol{c}_{p}\left(t\right) =\boldsymbol{W}_{c}\left(t\right)[\boldsymbol{\pi}_{p-1}\left(t\right);\mathcal{R}\left(\boldsymbol{y}_{p}\left(t\right)\right)]+\boldsymbol{b}_{c}\left(t\right), && \label{eq:zc}\\
& \check{\boldsymbol{\pi}}_{p}\left(t\right) =\mathcal{Q}\left(\boldsymbol{c}_{p}\left(t\right)\right)\odot\boldsymbol{\pi}_{p-1}\left(t\right), && \label{eq:picheck}\\
& \boldsymbol{\tilde{\pi}}_{p}\left(t\right) =\boldsymbol{W}_{\tilde{\pi}}\left(t\right)[\check{\boldsymbol{\pi}}_{p}\left(t\right);\mathcal{R}\left(\boldsymbol{y}_{p}\left(t\right)\right)]+\boldsymbol{b}_{\tilde{\pi}}\left(t\right), && \label{eq:pibar}\\
& \boldsymbol{\pi}_{p}\left(t\right) =\left(1-\mathcal{Q}\left(\boldsymbol{z}_{p}\left(t\right)\right)\right)\odot\boldsymbol{\pi}_{p-1}\left(t\right) && \nonumber \\
& +\mathcal{Q}\left(\boldsymbol{z}_{p}\left(t\right)\right)\odot\left(2\mathcal{Q}\left(\boldsymbol{\tilde{\pi}}_{p}\left(t\right)\right)-\boldsymbol{1}\right), && \label{eq:pit}
\end{flalign}
where $\mathcal{Q}(\cdot)$ is an activation function applied element-wise, which is often chosen as the sigmoid function in traditional GRU networks. In this paper, we set $\mathcal{Q}$ to the probit function, i.e., $\mathcal{Q}(x)=\int_{-\infty}^{x}\mathcal{N}(t;0,1)\,\mathrm{d}t$, to facilitate message passing. The RGRU parameters are summarized in a set $\boldsymbol{\omega}(t)=\left\{ \boldsymbol{W}_{\tilde{y}}(t),\boldsymbol{W}_{z}(t),\boldsymbol{W}_{c}(t),\boldsymbol{W}_{\tilde{\pi}}(t),\boldsymbol{b}_{\tilde{y}}(t),\boldsymbol{b}_{z}(t),\boldsymbol{b}_{c}(t),\boldsymbol{b}_{\tilde{\pi}}(t)\right\}$ which are estimated via approximate message passing.

\subsection{Two-Timescale Markov Priors}
\label{subsec:markov-priors}
We adopt a random-process framework over time slots to jointly capture the angular-domain channel $\boldsymbol{x}$ and impairment-related parameters $\boldsymbol{\omega}$. Owing to small-scale fading, the channel is treated as approximately stationary within each slot, yet it varies rapidly across successive slots~\cite{zinielDynamic2013a}. The parameters of the memory hardware impairment model, on the other hand, are assumed to evolve slowly throughout the channel tracking process. Consequently, the channel dynamics are modeled by a fast-varying Markov chain across slots, while the RGRU weights follow a slowly-varying Gaussian Markov chain.

Specifically, for the channel $\boldsymbol{x}\left(t\right)$, we define two hidden variables $\bar{\boldsymbol{s}}\left(t\right)$ and $\bar{\boldsymbol{x}}\left(t\right)$, where $\bar{\boldsymbol{s}}\left(t\right)$ denotes the support vector to indicate activity, and $\bar{\boldsymbol{x}}\left(t\right)$ represents $\boldsymbol{x}\left(t\right)$ when active~\cite{lianExploiting2019}. Therefore, $\boldsymbol{x}\left(t\right)$ can be written as (\ref{eq:hadamard}):
\begin{equation}
\boldsymbol{x}\left(t\right)=\bar{\boldsymbol{s}}\left(t\right)\odot\bar{\boldsymbol{x}}\left(t\right),\label{eq:hadamard}
\end{equation}
where $\bar{\boldsymbol{s}}\left(t\right)$ represents the binary support vector, which describes the dynamic sparsity of the channel support pattern, and the complex-valued vector $\bar{\boldsymbol{x}}\left(t\right)$ represents the hidden value. We assume independent dynamic priors for $\bar{\boldsymbol{s}}\left(t\right)$ and $\bar{\boldsymbol{x}}\left(t\right)$, i.e.:
\[
p\left(\bar{\boldsymbol{s}}\left(t\right)|\bar{\boldsymbol{s}}\left(t-1\right)\right)=\prod_{n=1}^{N}p\left(\bar{s}_{n}\left(t\right)|\bar{s}_{n}\left(t-1\right)\right),
\]
\[
p\left(\bar{\boldsymbol{x}}\left(t\right)|\bar{\boldsymbol{x}}\left(t-1\right)\right)=\prod_{n=1}^{N}p\left(\bar{x}_{n}\left(t\right)|\bar{x}_{n}\left(t-1\right)\right),
\]
where
\begin{equation}
\begin{aligned}
& p\left(\bar{s}_{n}\left(t\right)|\bar{s}_{n}\left(t-1\right)\right) \\
& =\begin{cases}
\rho_{01}\delta\left(\bar{s}_{n}\left(t\right)-1\right)+\rho_{00}\delta\left(\bar{s}_{n}\left(t\right)\right) & \bar{s}_{n}\left(t-1\right)=0\\
\rho_{11}\delta\left(\bar{s}_{n}\left(t\right)-1\right)+\rho_{10}\delta\left(\bar{s}_{n}\left(t\right)\right) & \bar{s}_{n}\left(t-1\right)=1
\end{cases},
\end{aligned}
\label{eq:dst}
\end{equation}
\begin{equation}
\begin{aligned}
& p\left(\bar{x}_{n}\left(t\right)|\bar{x}_{n}\left(t-1\right)\right) \\
& =\mathcal{CN}\left(\bar{x}_{n}\left(t\right);\left(1-\alpha\right)\bar{x}_{n}\left(t-1\right)+\alpha\xi,\alpha^{2}\kappa\right),
\end{aligned}
\label{eq:dxt}
\end{equation}
with $\rho_{00}=1-\rho_{01},\rho_{10}=1-\rho_{11}$, and we define $\bar{d}_{s,n}^{t}=p\left(\bar{s}_{n}\left(t\right)|\bar{s}_{n}\left(t-1\right)\right),\bar{d}_{x,n}^{t}=p\left(\bar{x}_{n}\left(t\right)|\bar{x}_{n}\left(t-1\right)\right)$ in the following parts. Note that $\rho_{01}/\rho_{11}$ represents the probability that the support turns active from inactive/active, $\alpha$ represents the temporal correlation between time slots, $\xi$ represents the mean of the angular-domain channel process, and $\kappa$ represents the variance of Gaussian perturbation. Larger $\rho_{01}$, $\rho_{10}$, $\alpha$, and $\kappa$ mean that the channel varies faster over time.

For the RGRU hidden states, we abbreviate the updating rules in (\ref{eq:zt})--(\ref{eq:pit}) at symbol $p$ to $\boldsymbol{\pi}_{p}(t)=\mathrm{GRU}(\boldsymbol{\pi}_{p-1}(t),\mathcal{R}(\boldsymbol{y}_{p}(t));\boldsymbol{\omega}(t))$ in order to construct the top-level factor graph. To track slow impairment drift, we model each element $\omega_i(t)\in\boldsymbol{\omega}(t)$ with a first-order Gaussian Markov transition,
\begin{equation}
p\left(\omega_{i}\left(t\right)|\omega_{i}\left(t-1\right)\right)=\mathcal{N}\left(\omega_{i}\left(t\right);\omega_{i}\left(t-1\right),\sigma_{\omega_i}^{2}\right),\label{eq:domegat}
\end{equation}
where $\sigma_{\omega_i}^{2}$ controls the per-slot drift variance of $\omega_i$, with $\sigma_{\omega_i}^{2}$ chosen much smaller than the channel transition variance so that $\boldsymbol{\omega}(t)$ evolves on a slower timescale than $\bar{\boldsymbol{s}}(t)$ and $\bar{\boldsymbol{x}}(t)$.

\section{Proposed Algorithm}
\label{sec:Joint-Estimation}
In this section, we detail the MP-TTBDL framework to jointly track the channel and the impairments from online pilots, based on the models in Section~II. MP-TTBDL consists of two components: inter-slot forward message passing, which leverages the two-timescale Markov priors to propagate temporal messages across slots, and per-slot iterative message passing, which performs joint Bayesian inference of the channel and impairment parameters. The two components are elaborated in the following subsections.

\subsection{Top-Level Factor Graph and Message Passing Overview}
To establish a Bayesian framework for joint channel estimation and impairment tracking, we formulate the probabilistic model over $T$ slots. The objective is to estimate the marginal posteriors of channel variables $\{\bar{\boldsymbol{s}}(t),\bar{\boldsymbol{x}}(t),\boldsymbol{x}(t)\}$ and impairment parameters $\{\boldsymbol{\omega}(t)\}$ from distorted observations $\{\tilde{\boldsymbol{y}}_{p}(t)\}_{p=1}^{P}$ for $t=1,\ldots,T$. Combining the RGRU-based observation model in (\ref{eq:resnet})--(\ref{eq:pit}), the two-timescale Markov prior in (\ref{eq:dst})--(\ref{eq:domegat}), and the channel observation model in (\ref{eq:obs}), the joint distribution of all random variables is written as:
\begin{flalign}
& p\Bigl(\bigl\{\bar{\boldsymbol{s}}(t),\bar{\boldsymbol{x}}(t),\boldsymbol{x}(t),\{\boldsymbol{y}_{p}(t),\boldsymbol{\pi}_{p}(t),\tilde{\boldsymbol{y}}_{p}(t)\}_{p=1}^{P},\boldsymbol{\omega}(t)\bigr\}_{t=1,\ldots,T}\Bigr) && \nonumber \\
& \propto p\left(\bar{\boldsymbol{s}}(1)\right)p\left(\bar{\boldsymbol{x}}(1)\right)p\left(\boldsymbol{\omega}(1)\right) && \nonumber \\
& \times \prod_{t=2}^{T}p\left(\bar{\boldsymbol{s}}(t)|\bar{\boldsymbol{s}}(t-1)\right)p\left(\bar{\boldsymbol{x}}(t)|\bar{\boldsymbol{x}}(t-1)\right)p\left(\boldsymbol{\omega}(t)|\boldsymbol{\omega}(t-1)\right) && \nonumber \\
& \times \prod_{t=1}^{T}\delta\left(\boldsymbol{x}(t)-\bar{\boldsymbol{s}}(t)\odot\bar{\boldsymbol{x}}(t)\right)\prod_{p=1}^{P}p\left(\boldsymbol{y}_{p}(t)|\boldsymbol{x}(t)\right) && \nonumber \\
& \times \prod_{t=1}^{T}\prod_{p=1}^{P}\delta\bigl(\boldsymbol{\pi}_{p}(t)-\mathrm{GRU}(\boldsymbol{\pi}_{p-1}(t),\mathcal{R}(\boldsymbol{y}_{p}(t));\boldsymbol{\omega}(t))\bigr) && \nonumber \\
& \times \prod_{t=1}^{T}p\left(\boldsymbol{\pi}_{0}(t)\right)\prod_{p=1}^{P}p\left(\tilde{\boldsymbol{y}}_{p}(t)|\boldsymbol{\pi}_{p}(t),\boldsymbol{y}_{p}(t),\boldsymbol{\omega}(t)\right). && \label{eq:toplevel_factorization}
\end{flalign}
where $p(\boldsymbol{y}_{p}(t)|\boldsymbol{x}(t))$ and $p(\tilde{\boldsymbol{y}}_{p}(t)|\boldsymbol{\pi}_{p}(t),\boldsymbol{y}_{p}(t),\boldsymbol{\omega}(t))$ come from (\ref{eq:obs}) and (\ref{eq:resnet}), respectively:
\begin{equation}
p\left(\boldsymbol{y}_{p}(t)|\boldsymbol{x}(t)\right) = \mathcal{CN}\bigl(\boldsymbol{y}_{p}(t);\boldsymbol{A}(\boldsymbol{\vartheta})\boldsymbol{x}(t), \sigma^{2}I_{N}\bigr).\label{eq:pyxsw}
\end{equation}
\begin{flalign}
& p\left(\tilde{\boldsymbol{y}}_{p}(t)|\boldsymbol{\pi}_{p}(t),\boldsymbol{y}_{p}(t),\boldsymbol{\omega}(t)\right) && \nonumber \\
& = \mathcal{N}\bigl(\mathcal{R}\left(\tilde{\boldsymbol{y}}_{p}(t)\right); \mathcal{R}\left(\boldsymbol{y}_{p}(t)\right)+\boldsymbol{W}_{\tilde{y}}(t)\boldsymbol{\pi}_{p}(t)+\boldsymbol{b}_{\tilde{y}}(t),\tilde{\sigma}^{2}\boldsymbol{I}_{2N}\bigr). && \label{eq:likelihood}
\end{flalign}
In addition, $p(\boldsymbol{\pi}_{0}(t))=\mathcal{N}(\boldsymbol{\pi}_{0}(t);\boldsymbol{0}_{N_{\pi}},\sigma_{\pi_0}^{2}\boldsymbol{I}_{N_{\pi}})$ assigns a zero-mean prior with small variance to the initial RGRU hidden state at the first pilot of each slot.

\begin{figure*}[!t]
\centering
\includegraphics[width=\linewidth]{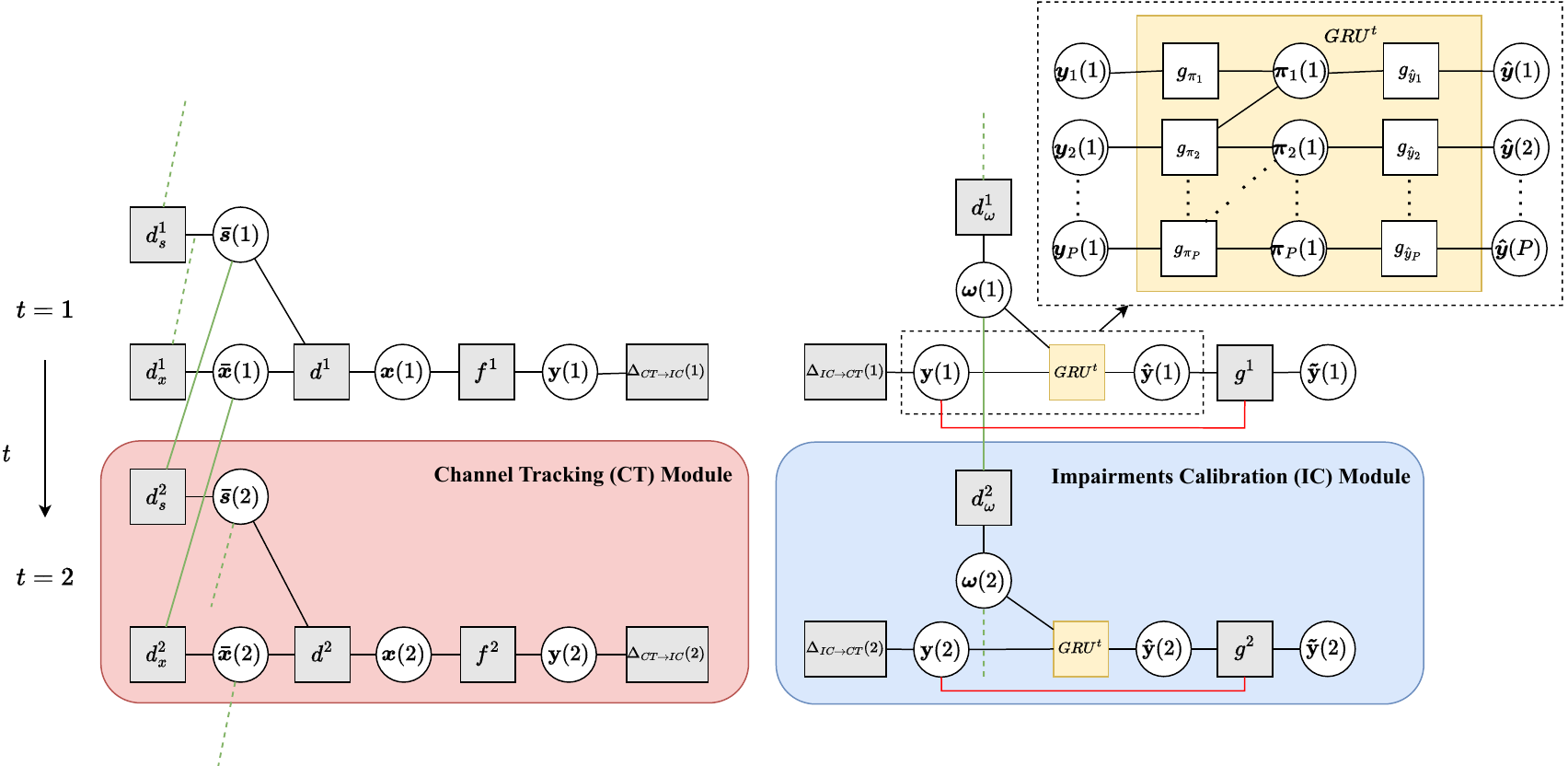}
\caption{Top-level factor graph for \eqref{eq:toplevel_factorization} with factor definitions in Table~\ref{tab:top-level-factors}.}
\label{fig:Top-level-factor}
\end{figure*}

To compute the marginal posterior distributions, we construct the multi-slot factor graph shown in Fig.~\ref{fig:Top-level-factor}, with the factor nodes summarized in Table~\ref{tab:top-level-factors}, and develop a corresponding approximate message passing algorithm. For notational convenience, we define the stacked observation vector $\mathbf{y} = [\boldsymbol{y}_1(t)^{\mathrm{T}},\ldots,\boldsymbol{y}_P(t)^{\mathrm{T}}]^{\mathrm{T}}\in\mathbb{C}^{NP}$ and the stacked noise vector $\mathbf{n}$ similarly, under which the per-slot observation model can be rewritten in the compact form $\mathbf{y} = (\boldsymbol{r}\otimes\boldsymbol{A}(\boldsymbol{\vartheta}))\,\boldsymbol{x} + \mathbf{n}$, where $\boldsymbol{r}$ collects the $P$ pilot symbols. At each slot $t$, inter-slot messages from slot $t-1$ are first combined with the two-timescale Markov transition factors to form equivalent priors on $\bar{\boldsymbol{s}}(t)$, $\bar{\boldsymbol{x}}(t)$, and $\boldsymbol{\omega}(t)$. Within the slot, the factor graph is partitioned into a channel tracking (CT) module and an impairments calibration (IC) module, which exchange extrinsic information over $\tau_{\max}$ turbo iterations: the CT module uses Turbo-OAMP to estimate the sparse channel $\boldsymbol{x}(t)$, while the IC module uses DAMP to update the RGRU parameters $\boldsymbol{\omega}(t)$. After convergence, the updated posteriors are forwarded to slot $t+1$ via inter-slot message passing, and the process repeats. Meanwhile, based on the channel estimation results from the CT module, an EM-based surrogate function for the angular-domain grid can be constructed, and the grid points are dynamically updated via gradient descent as detailed in~\cite{lianExploiting2019}. We shall use $\Delta_{a\rightarrow b}$ to denote the message from node $a$ to node $b$.

\begin{table}[!t]
\raggedright
\captionsetup{justification=raggedright,singlelinecheck=false}
\footnotesize
\setlength{\tabcolsep}{3pt}
\caption{Illustration of factor nodes in Fig.~\ref{fig:Top-level-factor}.}
\label{tab:top-level-factors}
\noindent
\begin{tblr}{
  width = \columnwidth,
  colspec = {Q[l,t,1.75] X[l,t,3.55] X[l,t,3.7]},
  hlines,
  vlines,
  rowsep = 2pt,
}
Factor & Distribution & Functional form \\
$d_{s}^{t}$ & $p(\bar{\boldsymbol{s}}(t)|\bar{\boldsymbol{s}}(t-1))$ & (\ref{eq:dst}) \\
$d_{x}^{t}$ & $p(\bar{\boldsymbol{x}}(t)|\bar{\boldsymbol{x}}(t-1))$ & (\ref{eq:dxt}) \\
$d^{t}$ & $p(\boldsymbol{x}(t)|\bar{\boldsymbol{s}}(t),\bar{\boldsymbol{x}}(t))$ & $\delta(\boldsymbol{x}(t)-\bar{\boldsymbol{s}}(t)\odot\bar{\boldsymbol{x}}(t))$ \\
$f^{t}$ & $p(\mathbf{y}(t)|\boldsymbol{x}(t))$ & (\ref{eq:pyxsw}) \\
$\Delta_{CT\rightarrow IC}(t)$ & Extrinsic message from CT to IC on $\mathbf{y}(t)$ & (\ref{deltaCT2IC}) \\
$d_{\omega}^{t}$ & $p(\boldsymbol{\omega}(t)|\boldsymbol{\omega}(t-1))$ & (\ref{eq:domegat}) \\
$\Delta_{IC\rightarrow CT}(t)$ & Extrinsic message from IC to CT on $\mathbf{y}(t)$ & (\ref{eq:IC2CT}) \\
$GRU^{t}$ & $p(\hat{\mathbf{y}}(t)|\mathbf{y}(t),\boldsymbol{\omega}(t))$ & \resizebox{\linewidth}{!}{$\delta(\hat{\mathbf{y}}(t)-\mathrm{GRU}(\mathbf{y}(t);\boldsymbol{\omega}(t)))$} \\
$g^{t}$ & \resizebox{\linewidth}{!}{$p(\tilde{\mathbf{y}}(t)|\hat{\mathbf{y}}(t),\mathbf{y}(t),\boldsymbol{\omega}(t))$} & (\ref{eq:likelihood}) \\
\end{tblr}
\end{table}

\subsection{Forward Message Passing Across Slots}
\label{subsec:forward-across-slots}
Here we only consider forward message passing across time slots. Based on sum-product rule, the messages are computed as follows:
\begin{equation}
\Delta_{\bar{\boldsymbol{s}}\left(t\right)\rightarrow d_{s}^{t+1}}=\Delta_{d_{s}^{t}\rightarrow\bar{\boldsymbol{s}}\left(t\right)}\cdot\Delta_{d^{t}\rightarrow\bar{\boldsymbol{s}}\left(t\right)},\label{eq:forwards}
\end{equation}
\begin{equation}
\Delta_{\bar{\boldsymbol{x}}\left(t\right)\rightarrow d_{x}^{t+1}}=\Delta_{d_{x}^{t}\rightarrow\bar{\boldsymbol{x}}\left(t\right)}\cdot\Delta_{d^{t}\rightarrow\bar{\boldsymbol{x}}\left(t\right)},\label{eq:forwardx}
\end{equation}
\begin{equation}
\Delta_{\boldsymbol{\omega}(t)\rightarrow d_{\omega}^{t+1}}=\Delta_{d_{\omega}^{t}\rightarrow\boldsymbol{\omega}(t)}\cdot\Delta_{GRU^{t}\rightarrow\boldsymbol{\omega}(t)},\label{eq:forwardomg}
\end{equation}

For messages within the $t$-th time slot, we first denote equivalent input prior factors $\check{d}_{s}^{t},\check{d}_{x}^{t}$ and $\check{d}_{\omega}^{t}$ that respectively represent the prior information of $\bar{\boldsymbol{s}}(t)$, $\bar{\boldsymbol{x}}(t)$ and $\boldsymbol{\omega}(t)$ extracted from all previous observations:
\begin{equation}
\check{d}_{s}^{t}=\Delta_{d_{s}^{t}\rightarrow\bar{\boldsymbol{s}}\left(t\right)}=\int_{\bar{\boldsymbol{s}}\left(t-1\right)}p\left(\bar{\boldsymbol{s}}\left(t\right)|\bar{\boldsymbol{s}}\left(t-1\right)\right)\Delta_{\bar{\boldsymbol{s}}\left(t-1\right)\rightarrow d_{s}^{t}},\label{eq:dcheckst}
\end{equation}
\begin{equation}
\check{d}_{x}^{t}=\Delta_{d_{x}^{t}\rightarrow\bar{\boldsymbol{x}}\left(t\right)}=\int_{\bar{\boldsymbol{x}}\left(t-1\right)}p\left(\bar{\boldsymbol{x}}\left(t\right)|\bar{\boldsymbol{x}}\left(t-1\right)\right)\Delta_{\bar{\boldsymbol{x}}\left(t-1\right)\rightarrow d_{x}^{t}},\label{eq:dcheckxt}
\end{equation}
\begin{equation}
\check{d}_{\omega}^{t}=\Delta_{d_{\omega}^{t}\rightarrow\boldsymbol{\omega}(t)}=\int_{\boldsymbol{\omega}(t-1)}p\left(\boldsymbol{\omega}(t)|\boldsymbol{\omega}(t-1)\right)\Delta_{\boldsymbol{\omega}(t-1)\rightarrow d_{\omega}^{t}}.\label{eq:dcheckomegat}
\end{equation}
Based on the Markov priors in (\ref{eq:dst})--(\ref{eq:domegat}), (\ref{eq:dcheckst})--(\ref{eq:dcheckomegat}) can be evaluated in closed form.

\subsection{Message Passing Within Each Slot}
\label{subsec:within-slot}
At each time slot $t$, the receiver performs joint inference by iteratively exchanging messages between the CT and IC modules for $\tau_{\max}$ turbo iterations. In each turbo iteration, the CT module updates messages via Turbo-OAMP, while the IC module updates messages via DAMP. Then, the two modules exchange extrinsic messages. In particular, $\Delta_{IC\rightarrow CT}$ aggregates the IC backward Gaussian messages on $\{\boldsymbol{y}_{p}(t)\}_{p=1}^{P}$ and is treated as equivalent AWGN side information in the CT module, whereas $\Delta_{CT\rightarrow IC}$ is formed via EP and is treated as equivalent AWGN side information in the IC module. After convergence, the messages passed to time slot $t+1$ are updated based on the sum-product rule as (\ref{eq:forwards})--(\ref{eq:forwardomg}). The message passing procedures with respect to the CT and IC modules are elaborated in the following; the slot index $t$ is dropped for brevity, as no ambiguity arises.

\subsubsection{CT Module (Turbo-OAMP)}
\label{subsec:Channel-Estimator}
The CT module estimates the sparse angular-domain channel $\boldsymbol{x}(t)$ from the stacked observations $\mathbf{y}$ within slot $t$. We apply Turbo-OAMP~\cite{lianExploiting2019} to perform this estimation efficiently. Specifically, the CT module is further divided into sub-module A and sub-module B, which extract information from the observations and the sparse prior factors, respectively. Firstly, sub-module B treats the extrinsic message w.r.t.\ $\boldsymbol{x}$ from sub-module A as equivalent AWGN observations, computes the posterior distribution of $\boldsymbol{x}$ based on standard sum-product rule, and then updates the extrinsic message to sub-module A via EP. Next, sub-module A treats the extrinsic message w.r.t.\ $\boldsymbol{x}$ from sub-module B as Gaussian prior, computes the posterior for $\boldsymbol{x}$ via linear minimum mean square error (LMMSE), and then updates the extrinsic message to sub-module B via EP. The two sub-modules exchange extrinsic messages to fuse the observations and sparse prior information until convergence. The detailed updating rules for messages within the CT module are the same as Turbo-OAMP. Please refer to~\cite{liuDownlink2018,lianExploiting2019} for the details.

Here we only give the updating rule for the top-level extrinsic message $\Delta_{CT\rightarrow IC}$ from the CT module to the IC module. Plugging in the message $\Delta_{\boldsymbol{x}\rightarrow f}=\mathcal{CN}(\boldsymbol{x};\boldsymbol{\mu}_{\boldsymbol{x}\rightarrow f},\varSigma_{\boldsymbol{x}\rightarrow f})$ computed via Turbo-OAMP, the mean and variance of message $\Delta_{f\rightarrow\boldsymbol{y}_{p}}$ for each symbol $p$ are given by:
\[
\boldsymbol{\mu}_{f\rightarrow\boldsymbol{y}_{p}}=\boldsymbol{A}(\boldsymbol{\vartheta})\boldsymbol{\mu}_{\boldsymbol{x}\rightarrow f},\quad \varSigma_{f\rightarrow\boldsymbol{y}_{p}}=\boldsymbol{A}(\boldsymbol{\vartheta})\varSigma_{\boldsymbol{x}\rightarrow f}\boldsymbol{A}^{H}(\boldsymbol{\vartheta})+\sigma^{2}I_{N}.
\]
The covariance matrix is not diagonal, and thus the extrinsic message is further approximated via EP:
\begin{equation}
\Delta_{CT\rightarrow IC} \approx\frac{\mathcal{P}\left\{ \Delta_{IC\rightarrow CT}\prod_{p=1}^{P}\Delta_{f\rightarrow\boldsymbol{y}_{p}}\right\} }{\Delta_{IC\rightarrow CT}}.
\label{deltaCT2IC}
\end{equation}
where $\mathcal{P}\left\{ \cdot\right\} $ approximates the distribution as a product of independent Gaussian via moment matching.

\subsubsection{IC Module (DAMP for RGRU)}
\label{subsec:ICModule}

\begin{figure}[!t]
\centering
\includegraphics[width=\linewidth]{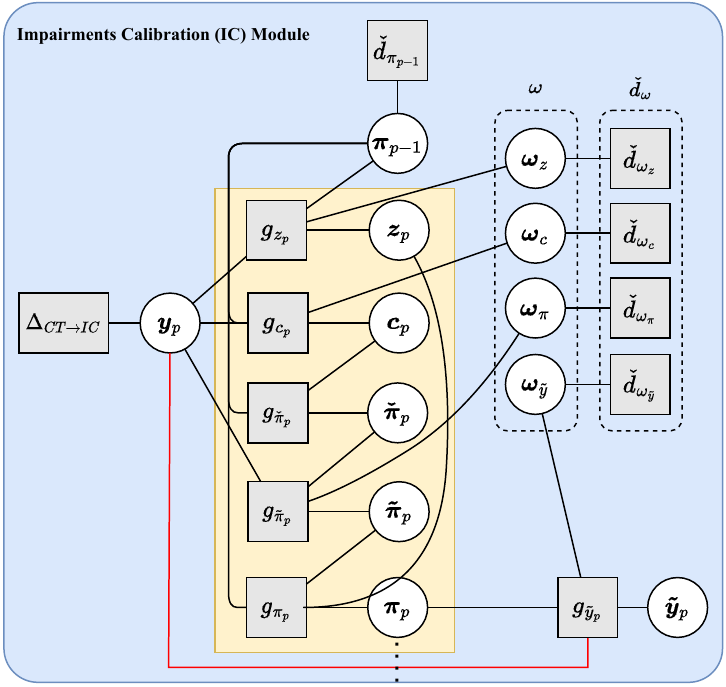}
\caption{Factor graph of the RGRU-based IC module at symbol $p$.}
\label{fig:ICmodule}
\end{figure}

Under the turbo framework, the IC module within each slot is separated for processing the messages associated with the RGRU-based impairment model, where the memory hardware impairments evolve over the $P$ pilot symbols. Based on the observation model given in \eqref{eq:resnet}--\eqref{eq:pit}, we take the processing of the $p$-th symbol as an illustrative example and construct the detailed factor graph of the IC module, as depicted in Fig.~\ref{fig:ICmodule}, with the definitions of the factor nodes provided in Table~\ref{tab:IC-boundary-messages}. Owing to the recurrent state transitions and gating nonlinearities inherent in the RGRU architecture, the resulting factor graph contains numerous loops, making exact marginal inference computationally intractable. While the DAMP framework in~\cite{xuBayesian2024} provides a powerful tool for Bayesian learning in feedforward deep neural networks (DNNs), it cannot be directly applied to our recurrent architecture. Extending DAMP to the RGRU presents two fundamental design challenges: (i) The recurrent state transition creates a non-trivial, loopy factor graph that requires a carefully coordinated message schedule, unlike the straightforward layer-wise schedule in feedforward networks. (ii) Both the probit activation functions $\mathcal{Q}(\cdot)$ and the element-wise products between gates and states introduce non-Gaussian messages, requiring tailored approximations to keep the message passing tractable. To address these challenges, we develop a custom DAMP procedure tailored to the RGRU. Our solution establishes a principled intra-symbol forward message pass (FMP) and backward message pass (BMP) schedule, as summarized in Table~\ref{tab:damp-schedule}, that systematically propagates messages along the $P$ pilot symbols while respecting the complex loopy structure of the recurrent graph. Furthermore, for the nonlinear submodules, we derive closed-form variational Bayesian inference (VBI) updates in Appendix~\ref{app:nonlinear-vbi} to approximate the otherwise intractable messages. This extension of the DAMP framework to a recurrent, gated architecture is a key methodological contribution that enables efficient online Bayesian learning of the IC module. After completing message passing, the extrinsic message $\Delta_{IC\to CT}$ is approximated as the product of the backward Gaussian messages over all symbols, which is fed to the CT module as equivalent observations. The entire IC module naturally decomposes into four bilinear submodules and two nonlinear submodules, each with its own equivalent input priors and observation model. In the following, we name each submodule by its output and detail the message passing rules.

\begin{table}[!t]
\raggedright
\captionsetup{justification=raggedright,singlelinecheck=false}
\footnotesize
\setlength{\tabcolsep}{3pt}
\caption{Illustration of factor nodes in Fig.~\ref{fig:ICmodule}.}
\label{tab:IC-boundary-messages}
\noindent
\begin{tblr}{
  width = \columnwidth,
  colspec = {Q[l,m,1.9] X[l,m,5.0] X[l,m,2.1]},
  hlines,
  vlines,
  rowsep = 2pt,
}
Factor & Distribution & Functional form \\
$\Delta_{CT\rightarrow IC}$ & Extrinsic message from CT to IC on $\boldsymbol{y}_{p}$ & (\ref{deltaCT2IC}) \\
$\check{d}_{\pi_{p-1}}$ & Equivalent prior for $\boldsymbol{\pi}_{p-1}$ & (\ref{eq:pi_prior}) \\
$\check{d}_{\omega_{z}}$ & Equivalent prior for $\boldsymbol{\omega}_{z}$ & \SetCell[r=4,c=1]{m} (\ref{eq:dcheckomegat}) \\
$\check{d}_{\omega_{c}}$ & Equivalent prior for $\boldsymbol{\omega}_{c}$ & \\
$\check{d}_{\omega_{\pi}}$ & Equivalent prior for $\boldsymbol{\omega}_{\pi}$ & \\
$\check{d}_{\omega_{\tilde{y}}}$ & Equivalent prior for $\boldsymbol{\omega}_{\tilde{y}}$ & \\
$g_{z_{p}}$ & $p(\boldsymbol{z}_{p}|\boldsymbol{\pi}_{p-1},\boldsymbol{y}_{p},\boldsymbol{\omega}_{z})$ & \SetCell[r=6,c=1]{m,l} $\delta$ function from observation models (\ref{eq:zt})--(\ref{eq:pit}) and (\ref{eq:likelihood}), respectively. \\
$g_{c_{p}}$ & $p(\boldsymbol{c}_{p}|\boldsymbol{\pi}_{p-1},\boldsymbol{y}_{p},\boldsymbol{\omega}_{c})$ & \\
$g_{\check{\pi}_{p}}$ & $p(\check{\boldsymbol{\pi}}_{p}|\boldsymbol{c}_{p},\boldsymbol{\pi}_{p-1})$ & \\
$g_{\tilde{\pi}_{p}}$ & $p(\tilde{\boldsymbol{\pi}}_{p}|\check{\boldsymbol{\pi}}_{p},\boldsymbol{y}_{p},\boldsymbol{\omega}_{\pi})$ & \\
$g_{\pi_{p}}$ & $p(\boldsymbol{\pi}_{p}|\boldsymbol{z}_{p},\boldsymbol{\pi}_{p-1},\tilde{\boldsymbol{\pi}}_{p})$ & \\
$g_{\tilde{y}_{p}}$ & $p(\mathcal{R}(\tilde{\boldsymbol{y}}_{p})|\boldsymbol{\pi}_{p},\boldsymbol{y}_{p},\boldsymbol{\omega}_{\tilde{y}})$ & \\
\end{tblr}
\end{table}

Four bilinear submodules $\boldsymbol{z}$, $\boldsymbol{c}$, $\tilde{\boldsymbol{\pi}}$, and $\tilde{\boldsymbol{y}}$, share the same equivalent observation model:
\begin{equation}
\hat{\boldsymbol{y}}_{\mathrm{eq}}=\boldsymbol{W}_{Y}\boldsymbol{u}_{\mathrm{eq}}+\boldsymbol{b}_{Y}+\boldsymbol{n}_{\mathrm{eq}},
\end{equation}
where $\boldsymbol{W}_{Y}$, $\boldsymbol{u}_{\mathrm{eq}}$, and $\boldsymbol{b}_{Y}$ are jointly estimated, the entries of $\boldsymbol{n}_{\mathrm{eq}}$ are i.i.d.\ AWGN.
At symbol $p$ of slot $t$, their corresponding equivalent inputs are $[\boldsymbol{\pi}_{p-1};\mathcal{R}(\boldsymbol{y}_{p})]$, $[\boldsymbol{\pi}_{p-1};\mathcal{R}(\boldsymbol{y}_{p})]$, $[\check{\boldsymbol{\pi}}_{p};\mathcal{R}(\boldsymbol{y}_{p})]$, and $\boldsymbol{\pi}_{p}$, respectively.
Since each bilinear submodule is identical to the linear layer model in the DAMP framework of~\cite{xuBayesian2024}. The forward and backward message updates are similar to Bayesian generalized approximate message passing (BiGAMP) in~\cite{parkerBiGAMP2014} (refer to~\cite{xuBayesian2024} for details).

Two submodules, $\check{\boldsymbol{\pi}}$ and $\boldsymbol{\pi}$, deal with observation models
\begin{equation}
\hat{\boldsymbol{y}}_{\mathrm{eq}}=f(\boldsymbol{u}_{\mathrm{eq}})+\boldsymbol{n}_{\mathrm{eq}},
\end{equation}
where $f(\cdot)$ denotes a nonlinear transformation and $\boldsymbol{u}_{\mathrm{eq}}$ collects the submodule input variables.
Specifically, in submodule $\check{\boldsymbol{\pi}}$, $f$ is defined by (\ref{eq:picheck}), $\boldsymbol{u}_{\mathrm{eq}}=\{\boldsymbol{c}_{p},\boldsymbol{\pi}_{p-1}\}$, and the equivalent observation refers to the Gaussian backward message $\Delta_{\check{\boldsymbol{\pi}}\rightarrow g_{\check{\boldsymbol{\pi}}}^{t}}$.
In submodule $\boldsymbol{\pi}$, $f$ is defined by (\ref{eq:pit}), $\boldsymbol{u}_{\mathrm{eq}}=\{\boldsymbol{z}_{p},\boldsymbol{\pi}_{p-1},\tilde{\boldsymbol{\pi}}_{p}\}$, and the equivalent observation refers to the Gaussian backward message $\Delta_{\boldsymbol{\pi}\rightarrow g_{\boldsymbol{\pi}}^{t}}$.
Because the observation models involve nonlinear mappings through $\mathcal{Q}(\cdot)$, some message computations are intractable; we therefore introduce mean-field VBI. We take submodule $\boldsymbol{\pi}$ as an example to describe the computation as detailed in Appendix~\ref{app:nonlinear-vbi}.
\label{subsec:nonlinear-vbi}

\label{subsec:bilinear-updates}
The detailed updating rules for the submodules are omitted for brevity. Each intra-symbol DAMP iteration consists of an FMP and a BMP: the FMP computes equivalent priors for the submodules sequentially in the order listed in Table~\ref{tab:damp-schedule}, while the BMP computes equivalent observations in the reverse order. Messages among submodules are combined via the sum-product rule, and non-Gaussian extrinsic messages are approximated by EP.

After the above message passing for symbol $p$ completes, the messages on the hidden state $\boldsymbol{\pi}_p$ are aggregated to form the equivalent prior for the message computation of the next symbol, i.e.,
\begin{equation}
\check{d}_{\pi_p} = \Delta_{g_{\pi_p} \to \boldsymbol{\pi}_p} \; \Delta_{g_{\tilde{y}_p} \to \boldsymbol{\pi}_p},
\label{eq:pi_prior}
\end{equation}
where the right-hand side collects the messages from the state transition factor $g_{\pi_p}$ and the output factor $g_{\tilde{y}_p}$ that point to $\boldsymbol{\pi}_p$. The entire forward message-passing procedure is executed sequentially for $p=1,\ldots,P$, propagating the accumulated impairment memory along the pilot chain within slot $t$.

\begin{table}[!t]
\raggedright
\captionsetup{justification=raggedright,singlelinecheck=false}
\footnotesize
\caption{Equivalent priors and observations for IC submodules.}
\label{tab:damp-schedule}
\noindent
\begin{tblr}{
  width = \columnwidth,
  colspec = {Q[l,m,1.65] X[l,t,4.5] X[l,t,2.4]},
  hlines,
  vlines,
}
Submodule & Prior & Observation \\
$\boldsymbol{z}$ & $\Delta_{\boldsymbol{y}_p\rightarrow g_{\boldsymbol{z}}^{t}}\,\Delta_{\boldsymbol{\pi}_{p-1}\rightarrow g_{\boldsymbol{z}}^{t}}\,\Delta_{\boldsymbol{\omega}_{z}\rightarrow g_{\boldsymbol{z}}^{t}}$ & $\Delta_{\boldsymbol{z}_p\rightarrow g_{\boldsymbol{z}}^{t}}$ \\
$\boldsymbol{c}$ & $\Delta_{\boldsymbol{y}_p\rightarrow g_{\boldsymbol{c}}^{t}}\,\Delta_{\boldsymbol{\pi}_{p-1}\rightarrow g_{\boldsymbol{c}}^{t}}\,\Delta_{\boldsymbol{\omega}_{c}\rightarrow g_{\boldsymbol{c}}^{t}}$ & $\Delta_{\boldsymbol{c}_p\rightarrow g_{\boldsymbol{c}}^{t}}$ \\
$\check{\boldsymbol{\pi}}$ & $\Delta_{\boldsymbol{c}_p\rightarrow g_{\check{\boldsymbol{\pi}}}^{t}}\,\Delta_{\boldsymbol{\pi}_{p-1}\rightarrow g_{\check{\boldsymbol{\pi}}}^{t}}$ & $\Delta_{\check{\boldsymbol{\pi}}_p\rightarrow g_{\check{\boldsymbol{\pi}}}^{t}}$ \\
$\tilde{\boldsymbol{\pi}}$ & $\Delta_{\check{\boldsymbol{\pi}}_p\rightarrow g_{\tilde{\boldsymbol{\pi}}}^{t}}\,\Delta_{\boldsymbol{y}_p\rightarrow g_{\tilde{\boldsymbol{\pi}}}^{t}}$ & $\Delta_{\tilde{\boldsymbol{\pi}}_p\rightarrow g_{\tilde{\boldsymbol{\pi}}}^{t}}$ \\
$\boldsymbol{\pi}$ & $\Delta_{\boldsymbol{z}_p\rightarrow g_{\boldsymbol{\pi}}^{t}}\,\Delta_{\boldsymbol{\pi}_{p-1}\rightarrow g_{\boldsymbol{\pi}}^{t}}\,\Delta_{\tilde{\boldsymbol{\pi}}_p\rightarrow g_{\boldsymbol{\pi}}^{t}}$ & $\Delta_{\boldsymbol{\pi}_p\rightarrow g_{\boldsymbol{\pi}}^{t}}$ \\
$\tilde{\boldsymbol{y}}$ & $\Delta_{\boldsymbol{y}_p\rightarrow g_{\tilde{\boldsymbol{y}}}^{t}}\,\Delta_{\boldsymbol{\pi}_p\rightarrow g_{\tilde{\boldsymbol{y}}}^{t}}\,\Delta_{\boldsymbol{\omega}_{\tilde{y}}\rightarrow g_{\tilde{\boldsymbol{y}}}^{t}}$ & $\Delta_{\mathcal{R}(\tilde{\boldsymbol{y}}_p)\rightarrow g_{\tilde{\boldsymbol{y}}}^{t}}$ \\
\end{tblr}
\end{table}

After message passing within the IC module, the output extrinsic message on $\{\boldsymbol{y}_{p}\}_{p=1}^{P}$ is updated by aggregating backward messages based on the sum-product rule:
\begin{equation}
\begin{split}
\Delta_{IC\to CT} &= \prod_{p=1}^{P}\Delta_{g_{\boldsymbol{z},p}^{t}\rightarrow\boldsymbol{y}_{p}}\,\Delta_{g_{\boldsymbol{c},p}^{t}\rightarrow\boldsymbol{y}_{p}} \\
&\times \Delta_{g_{\tilde{\boldsymbol{\pi}},p}^{t}\rightarrow\boldsymbol{y}_{p}}\,\Delta_{g_{\tilde{\boldsymbol{y}},p}^{t}\rightarrow\boldsymbol{y}_{p}},
\end{split}
\label{eq:IC2CT}
\end{equation}
which is sent to the CT module as the estimated observation without impairments.

The MP-TTBDL algorithm is summarized in Alg.~\ref{algorithm summary}.

\begin{algorithm}[!t]
\caption{Online joint channel and impairments tracking algorithm}
\label{algorithm summary}
\begin{algorithmic}[1]
\FOR{$t=1,2,\ldots$}
\STATE Compute input prior messages $\check{d}_{s}^{t},\check{d}_{x}^{t},\check{d}_{\omega}^{t}$. (Initialization if $t=1$)
\STATE Set $\Delta_{IC\rightarrow CT}$ to a non-informative (flat) Gaussian extrinsic message for the CT observation model.
\FOR{$\tau=1 \ldots \tau_{\max}$}
\STATE $\bullet$ CT Module:
\STATE Update messages within CT Module via Turbo-OAMP with input prior $\check{d}_{s}^{t},\check{d}_{x}^{t}$ and extrinsic message $\Delta_{IC\rightarrow CT}$.
\STATE Update output extrinsic message $\Delta_{CT\rightarrow IC}$ using (\ref{deltaCT2IC}).
\STATE $\bullet$ IC Module:
\STATE Update messages within IC Module via DAMP with input prior $\check{d}_{\omega}^{t}$ and extrinsic message $\Delta_{CT\rightarrow IC}$.
\STATE Update output extrinsic message $\Delta_{IC\rightarrow CT}$ using (\ref{eq:IC2CT}).
\ENDFOR
\STATE Update the output messages to the next time slot as (\ref{eq:forwards})--(\ref{eq:forwardomg}).
\STATE Update angular domain grid points based on gradient descent.
\ENDFOR
\end{algorithmic}
\end{algorithm}

\subsection{Complexity Analysis}
\label{subsec:complexity}

\begin{figure}[!t]
\centering
\includegraphics[width=\linewidth]{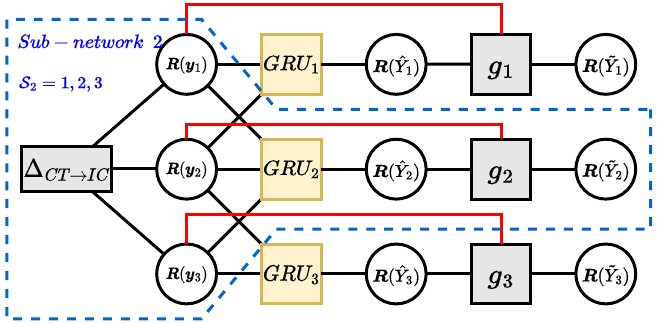}
\caption{Factor graph of the parallel sub-network structure, where weights and hidden states are omitted for simplicity.}
\label{fig:parallel-nn}
\end{figure}

To reduce model complexity, we exploit the localized crosstalk structure in practical massive MIMO receivers. Strong crosstalk is typically confined to adjacent RF chains~\cite{zhaoLowComplexity2023,uthayakumarScalable2023}. Consequently, the distorted sample on the $n$-th RF chain depends only on $\boldsymbol{y}_{\mathcal{S}_n}$, where $\mathcal{S}_n$ is the index set of the RF chains that have strong crosstalk with the $n$-th chain. The RGRU can therefore be decomposed into $N$ parallel sub-networks $\mathrm{GRU}_{n}$, each with input $\mathcal{R}(\boldsymbol{y}_{\mathcal{S}_{n}})$ and output in $\mathbb{R}^{2}$, as conceptually illustrated in Fig.~\ref{fig:parallel-nn}, where the weight matrices and the hidden state $\boldsymbol{\pi}_{p}$ are omitted for simplicity. Equivalently, constraining $\boldsymbol{W}_{z}(t)$, $\boldsymbol{W}_{c}(t)$, $\boldsymbol{W}_{\tilde{\pi}}(t)$, and $\boldsymbol{W}_{\tilde{y}}(t)$ to be block diagonal reduces the model parameters from $\mathcal{O}(N^2 N_{\mathrm{sub}}^2)$ to $\mathcal{O}(N N_{\mathrm{sub}}^2)$ without loss of fitting performance. In this parallel arrangement, each sub-network $\mathrm{GRU}_n$ maintains its own hidden state; let $N_{\mathrm{sub}}$ denote its sub-network hidden dimension. The collection of these local states constitutes the overall memory of the impairment model.

We compare the model sizes of the proposed RGRU with the two conventional compensators evaluated in Sec.~\ref{sec:simulation}: a GMP compensator and a GRU compensator that adopts the same parallel per-chain sub-network architecture as the RGRU. From (\ref{eq:resnet}) and (\ref{eq:zt})--(\ref{eq:pibar}), the proposed RGRU satisfies $|\boldsymbol{\omega}| = \sum_{n=1}^{N}\bigl(3N_{\mathrm{sub}}^{2} + 6N_{\mathrm{sub}}|\mathcal{S}_{n}| + 5N_{\mathrm{sub}} + 2\bigr) \sim \mathcal{O}(N N_{\mathrm{sub}}^{2})$. The parallel GRU compensator reuses the same per-chain sub-network architecture and therefore has the same $|\boldsymbol{\omega}|$. For the GMP compensator, each chain builds GMP bases with memory taps $m=0,\ldots,M$ and odd nonlinearity orders $\{1,3,\ldots,K\}$, giving $N_{b,n}=(M+1)\lceil K/2\rceil\,(2|\mathcal{S}_{n}|-1)$ basis functions per chain; with widely-linear fitting, the real coefficient count is $|\boldsymbol{\beta}| = 4\sum_{n=1}^{N} N_{b,n} \sim \mathcal{O}(N K M)$.

We next compare the per-slot computational complexity of Alg.~1 with the GMP and GRU compensators. All schemes share the same Turbo-OAMP channel estimator, which requires $\tau_{\max}$ internal iterations between sub-modules A and B and contributes $\mathcal{O}(\tau_{\max} P N^{2})$ operations per slot. Under the parallel per-chain architecture in Fig.~\ref{fig:parallel-nn}, the proposed MP-TTBDL additionally executes the IC module (DAMP) in each of the $\tau_{\max}$ CT--IC turbo iterations, introducing an extra $\mathcal{O}(\tau_{\max} P N N_{\mathrm{sub}}^{2} + \tau_{\max} P N N_{\mathrm{sub}})$ per slot, plus $\mathcal{O}(N N_{\mathrm{sub}}^{2})$ for the cross-slot message updates, yielding $\mathcal{C}_{\mathrm{TTBDL}} = \mathcal{O}(\tau_{\max} P N^{2} + \tau_{\max} P N N_{\mathrm{sub}}^{2} + \tau_{\max} P N N_{\mathrm{sub}} + N N_{\mathrm{sub}}^{2})$. For the baselines, the GMP compensator applies a fixed polynomial transform costing $\mathcal{O}(P N K p_{\mathrm{md}})$ and the GRU compensator uses a feedforward pass of complexity $\mathcal{O}(P N N_{\mathrm{sub}}^{2} + P N N_{\mathrm{sub}})$, each followed by the same Turbo-OAMP stage, giving $\mathcal{C}_{\mathrm{GMP}} = \mathcal{O}(P N K p_{\mathrm{md}} + \tau_{\max} P N^{2})$ and $\mathcal{C}_{\mathrm{GRU}} = \mathcal{O}(P N N_{\mathrm{sub}}^{2} + P N N_{\mathrm{sub}} + \tau_{\max} P N^{2})$. As $N$ grows large while $P$, $\tau_{\max}$, and $N_{\mathrm{sub}}$ remain fixed, the per-slot complexity of all three schemes asymptotes to $\mathcal{O}(\tau_{\max} P N^{2})$.

\section{Performance Evaluation}
\label{sec:simulation}
In this section, we present simulation results to evaluate the performance of the proposed algorithm. We first conduct offline pretraining to demonstrate the superiority of our message-passing-based neural network training algorithm over conventional methods. We then consider the case where the online impairments exactly match the offline pretraining configuration, and verify the effectiveness of the proposed joint Bayesian deep learning framework. Finally, we evaluate the scenario where the online impairments drift slowly over time, confirming both the effectiveness of the two-timescale Markov transition model and the robustness of the MP-TTBDL algorithm design.

%
%

\subsection{Offline Pretraining}
The receiver is equipped with $N = 64$ antennas in the following simulations. To generate representative training and test data, we synthesize the distorted observations by cascading a linear crosstalk model, a nonlinear amplifier with memory, and a phase-only IQ imbalance model, all of which are widely adopted in the literature for hardware impairment simulation. Crosstalk is assumed only between adjacent RF chains~\cite{zhaoLowComplexity2023,uthayakumarScalable2023}, with linear coupling coefficient $\epsilon=0.1778$ ($-15$\,dB). The IQ imbalance is modeled as a phase mismatch $\phi=0.1778$ ($-15$\,dB). The amplifier nonlinearity follows a GMP with memory depth $p_{\mathrm{md}}=4$ and nonlinearity order $K=3$, whose coefficients are taken from a measured-data example in MATLAB~\cite{mathworksGMPFitting}. These specific parametric models are used only to generate evaluation data and are not assumed by the proposed method, which can handle more general hardware impairments.

We generate $10{,}000$ independent sequences of 10 symbols, randomly split into $5{,}000$ training and $5{,}000$ test samples. For each symbol, the entries of the ideal received vector $\boldsymbol{y}_{p}\in\mathbb{C}^{N}$ are independently drawn from $\mathcal{CN}(0,1)$.

\begin{figure}[!t]
\centering
\begin{minipage}[t]{0.48\linewidth}
\centering
\includegraphics[width=\linewidth, clip]{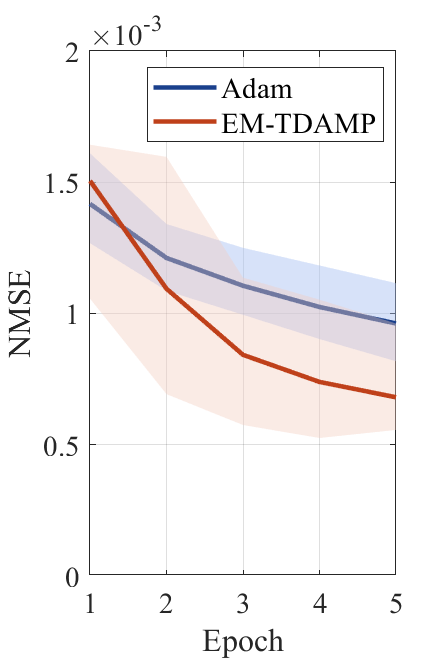}\\
\centerline{\footnotesize (a) $N_{\mathrm{sub}}=64$}
\end{minipage}\hfill
\begin{minipage}[t]{0.48\linewidth}
\centering
\includegraphics[width=\linewidth, clip]{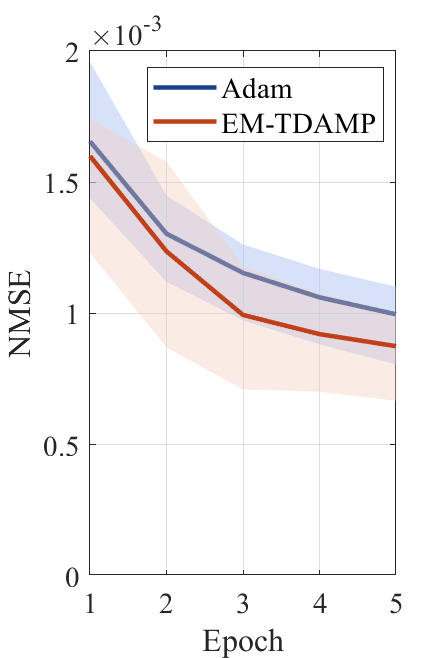}\\
\centerline{\footnotesize (b) $N_{\mathrm{sub}}=32$}
\end{minipage}
\caption{Fitting performance comparison of EM-TDAMP and Adam for $N_{\mathrm{sub}}=64$ and $N_{\mathrm{sub}}=32$.}
\label{fig:rf-test-nmse}
\end{figure}

To demonstrate the advantages of our expectation-maximization turbo deep approximate message passing (EM-TDAMP) algorithm~\cite{xuBayesian2024} in neural network training, we conducted simulations with two sub-network hidden dimensions, $N_{\mathrm{sub}}=32$ and $N_{\mathrm{sub}}=64$, for each sub-network shown in Fig.~\ref{fig:parallel-nn}. We used a batch size of $100$ and trained for $5$ epochs. For the Adam baseline~\cite{kingmaAdam2015}, the initial learning rate was set to $0.01$. The fitting performance is presented in Fig.~\ref{fig:rf-test-nmse}, which clearly shows that our algorithm achieves faster convergence and superior fitting accuracy than Adam under both configurations.

Our method performs Bayesian inference on the top-level factor graph during online tracking and requires no separately trained compensator. For fair comparison, we train two conventional compensators on the same pretraining dataset: a GMP compensator (nonlinearity order $K=3$, memory depth $p_{\mathrm{md}}=4$) with $15{,}168$ real coefficients obtained by least squares, and a GRU compensator that adopts the same parallel per-chain sub-network architecture with $N_{\mathrm{sub}}=32$, giving a total of $243{,}456$ real parameters identical to the proposed RGRU.

\subsection{Online Joint Tracking}
For online joint tracking, we used the same RGRU ensemble as in offline pretraining, with $N_{\mathrm{sub}}=32$ per sub-network. The time-varying flat-fading uplink channel $\boldsymbol{h}(t)$ was generated by the QuaDRiGa ray-tracing simulator~\cite{jaeckelQuaDRiGa2019} under the 3GPP TR~38.901 urban macrocell non-line-of-sight (NLOS) scenario, with carrier frequency $6$\,GHz, two propagation clusters, and per-cluster azimuth/elevation angular spreads of $1^\circ$. The user speed was set to $1$\,m/s, with $P=4$ pilot symbols per slot and a slot duration of $1$\,ms.

The proposed method is compared against the following benchmark schemes:
\begin{itemize}
\item \textbf{MP-BDL:} joint inference with a static $\boldsymbol{\omega}$ prior; used to validate the necessity of the slow Markov prior for tracking impairment drift.
\item \textbf{MP-BDL-Frozen:} CT--IC message passing with frozen $\boldsymbol{\omega}$; used to confirm the benefit of online impairment adaptation.
\item \textbf{w/o compensation:} Turbo-OAMP is applied directly to the distorted observations $\tilde{\boldsymbol{y}}$ to quantify the degradation caused by hardware impairments.
\item \textbf{GMP for compensation:} an offline-trained GMP compensator is followed by Turbo-OAMP to demonstrate the advantage over a conventional polynomial compensator.
\item \textbf{GRU for compensation:} an offline-trained parallel GRU compensator of the same model size is followed by Turbo-OAMP to assess the gain of the proposed joint Bayesian framework over a decoupled approach that performs compensation and channel estimation separately.
\end{itemize}

\begin{figure}[!t]
\centering
\includegraphics[width=0.95\linewidth, clip]{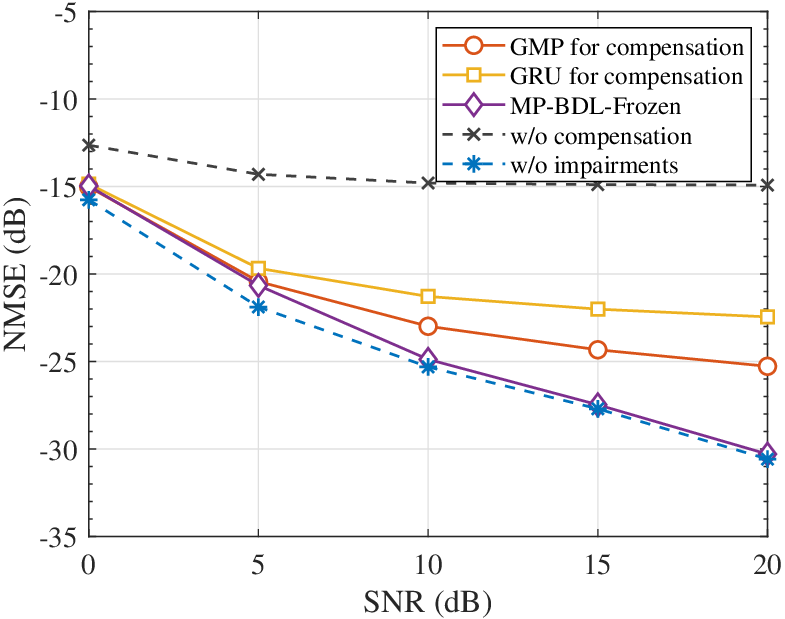}
\caption{Channel estimation NMSE versus SNR when the online impairments coincide with the offline pretraining.}
\label{fig:online-same-channel-snr}
\end{figure}

We first consider the scenario where the online impairments are identical to those used in offline pretraining. Since no impairment drift occurs, only MP-BDL-Frozen and the conventional compensators are compared, so as to validate the intra-slot joint Bayesian inference framework. The channel estimation NMSE is averaged over $100$ consecutive slots with $\tau_{\max}=5$ turbo iterations between the CT and IC modules. Fig.~\ref{fig:online-same-channel-snr} reports the channel NMSE versus SNR for all schemes, along with an upper bound obtained from impairment-free observations and a lower bound corresponding to the uncompensated case. Hardware impairments are seen to severely degrade channel estimation accuracy. Although conventional compensators mitigate this degradation to some extent, a considerable gap to the impairment-free upper bound persists. By contrast, the proposed MP-BDL-Frozen achieves the lowest channel NMSE among all compared schemes and closely approaches the upper bound across the entire SNR range, with the advantage being especially pronounced at high SNR.

\begin{figure}[!t]
\centering
\includegraphics[width=0.95\linewidth, clip]{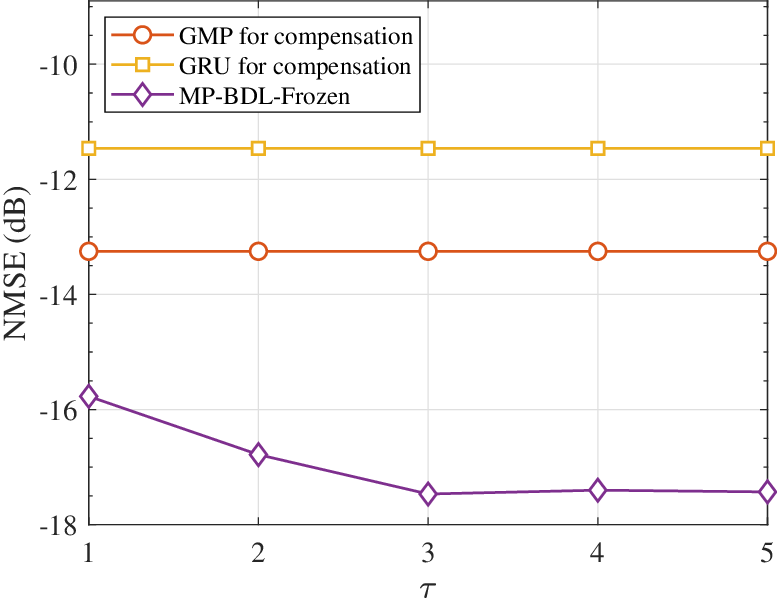}
\caption{NMSE of the estimated $\mathrm{y}$ versus $\tau$ at SNR$=20$\,dB when the online impairments coincide with the offline pretraining.}
\label{fig:online-same-rf-tau}
\end{figure}

To reveal the source of this gain, Fig.~\ref{fig:online-same-rf-tau} plots the NMSE of the recovered signal $\mathbf{y}$ versus the turbo iteration index $\tau$, where SNR$=20$\,dB is taken as a representative case. Under the joint Bayesian inference framework, the IC module reconstructs $\mathbf{y}$ by exploiting the prior information supplied by the CT module. As the turbo iterations proceed, the two modules progressively refine their extrinsic messages, yielding an estimate of $\mathbf{y}$ that is more accurate than that produced by the baseline compensators and, in turn, a lower channel estimation error. This gain becomes more prominent at high SNR, where the prior variance from the CT module is smaller and the prior information is therefore more precise, further amplifying the benefit of joint inference. It is observed that the NMSE converges within $\tau_{\max}=3$ iterations, and this value is consequently adopted as the default in all subsequent simulations.

The performance under slowly time-varying impairments is evaluated next, so as to validate the effectiveness of the proposed two-timescale Markov transition model MP-TTBDL algorithm. Let $\boldsymbol{\theta}(t)$ collect all time-varying impairment parameters at slot $t$, including the coefficients for LNA, the IQ phase mismatches, and the crosstalk coupling coefficients. At $t=0$, the crosstalk and IQ mismatch amplitudes are set to $\epsilon_n(0)=\phi_n(0)=0.2239$ ($-13$\,dB). The LNA coefficients are initialized by scaling the fixed coefficients used in the matched-online experiment by $1.02$, $1.05$, and $1.1$ for the linear, third-order, and fifth-order taps, respectively, yielding a degraded amplifier state. For $t=1,\ldots,600$, the parameters evolve according to $\boldsymbol{\theta}(t) = \rho\,\boldsymbol{\theta}(t-1) + (1-\rho)\,\boldsymbol{\theta}^{\star} + \eta\,|\boldsymbol{\theta}(t-1)|^{2}\odot\boldsymbol{\xi}(t)$, with $\rho=0.995$, $\eta=0.01$, and $\boldsymbol{\xi}(t)\sim\mathcal{N}(\boldsymbol{0},\boldsymbol{I})$ element-wise independent across slots. The target $\boldsymbol{\theta}^{\star}$ drives the IQ and crosstalk amplitudes from $-13$\,dB toward $-12$\,dB, and drives the third- and fifth-order LNA taps further upward by approximately $2\%$ and $5\%$ relative to their initial degraded values, while the linear LNA taps remain largely unchanged. Over the $600$ slots, the parameters therefore evolve gradually from the degraded initial condition toward the more severe impairment state specified by $\boldsymbol{\theta}^{\star}$, with the small per-slot step size governed by $\rho$ and $\eta$ capturing the slow drift characteristic of practical hardware aging and environmental variations.

\begin{figure}[!t]
\centering
\includegraphics[width=0.95\linewidth, clip]{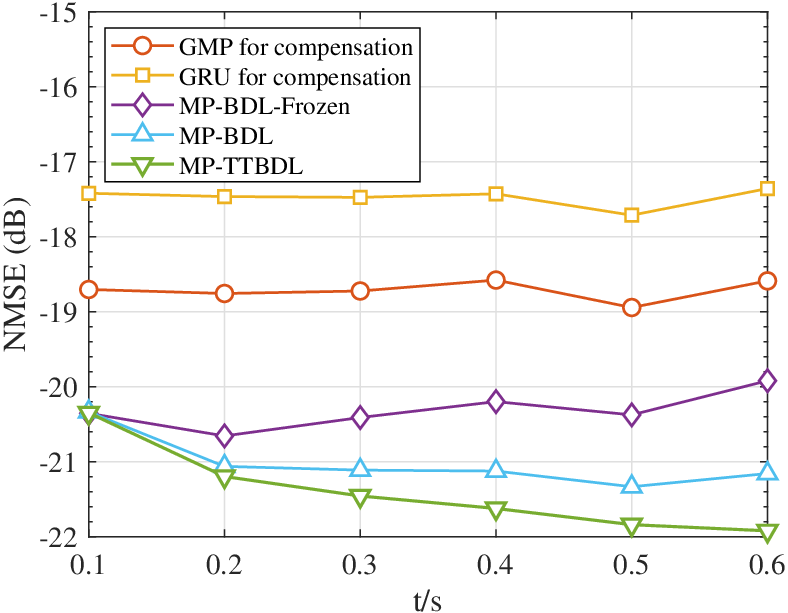}
\caption{Channel tracking NMSE versus $t$ under time-varying impairments at SNR$=20$\,dB, where each data point is averaged over $100$ slots ($0.1$\,s) and the total tracking duration is $600$ slots ($0.6$\,s).}
\label{fig:online-varying-channel}
\end{figure}

\begin{figure}[!t]
\centering
\includegraphics[width=0.95\linewidth, clip]{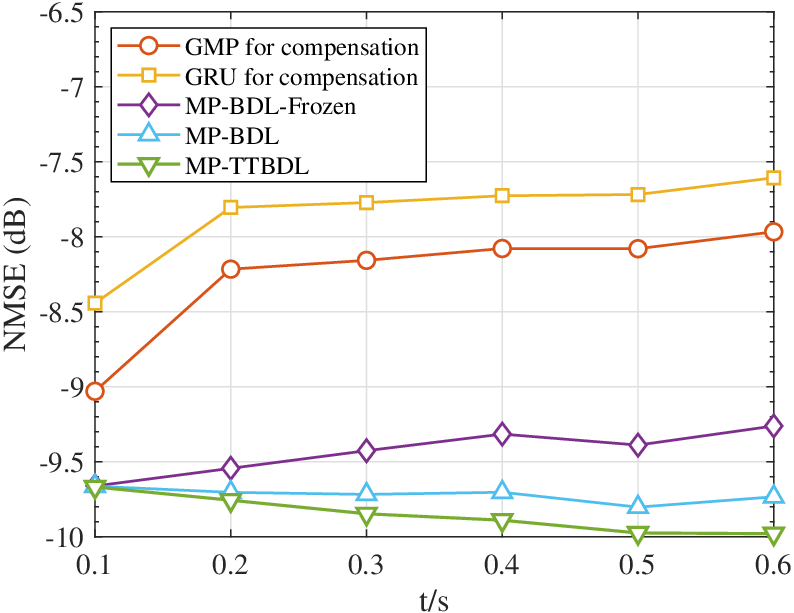}
\caption{NMSE of estimated $\mathbf{y}$ versus $t$.}
\label{fig:online-varying-rf}
\end{figure}

\begin{figure}[!t]
\centering
\includegraphics[width=0.95\linewidth, clip]{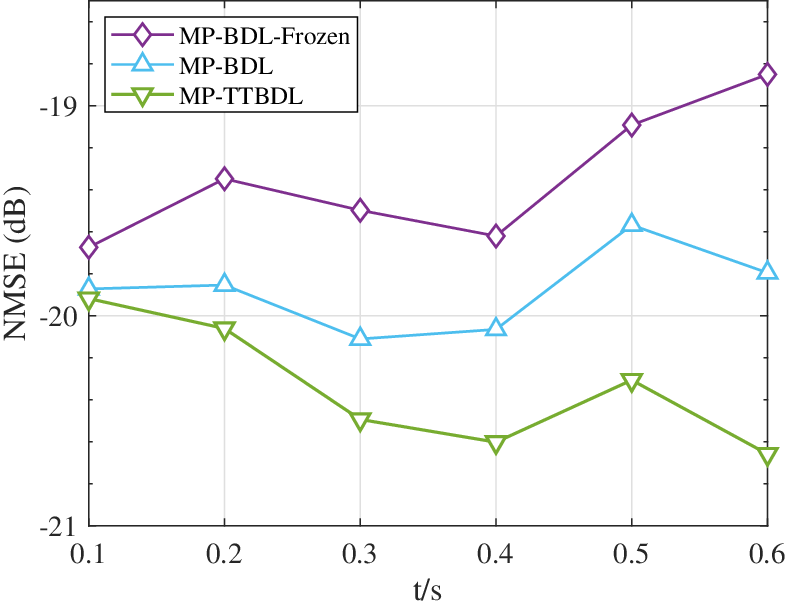}
\caption{Forward RGRU fitting NMSE versus $t$.}
\label{fig:online-varying-fitting-t}
\end{figure}

To evaluate the tracking capability under time-varying impairments, the channel estimation NMSE and the NMSE of the recovered signal $\boldsymbol{y}$ are first examined at SNR$=20$\,dB over $600$ slots. Fig.~\ref{fig:online-varying-channel} presents the channel estimation NMSE versus $t$. The MP-BDL-family algorithms (MP-BDL, MP-BDL-Frozen, and the proposed MP-TTBDL) consistently outperform the conventional schemes that apply compensation followed by Bayesian channel estimation. Within the MP-BDL family, MP-TTBDL achieves an increasingly larger gain over both MP-BDL and MP-BDL-Frozen as tracking proceeds, validating the effectiveness of the two-timescale Markov prior in refining the impairment parameter estimates over time. Fig.~\ref{fig:online-varying-rf} reports the NMSE of the recovered signal $\boldsymbol{y}$ versus $t$. It is observed that MP-TTBDL produces a more accurate estimate of $\boldsymbol{y}$ than the competing schemes, confirming that the channel estimation gain observed in Fig.~\ref{fig:online-varying-channel} originates from the improved recovery of the input signal $\boldsymbol{y}$, which in turn provides higher-quality input to the CT module. The forward RGRU fitting NMSE, defined as $\|\hat{\mathbf{y}}-\tilde{\mathbf{y}}\|_{\mathrm{F}}^{2}/\|\tilde{\mathbf{y}}\|_{\mathrm{F}}^{2}$, is further examined in Fig.~\ref{fig:online-varying-fitting-t} to isolate the contribution of the online RGRU parameter update. This metric quantifies how accurately the RGRU forward model reproduces the impaired observations. As online pilot data accumulate, MP-TTBDL progressively refines the RGRU parameters via inter-slot message passing, and the fitting NMSE exhibits an overall decreasing trend over time. MP-BDL, which retains the inter-slot message passing but lacks the slow Markov prior to model the gradual parameter drift, demonstrates a limited tracking capability and thus falls short of the proposed MP-TTBDL. The baselines without online adaptation, i.e., MP-BDL-Frozen and other baseline schemes, are unable to adjust the impairment model based on live observations and show no improvement over time. These results confirm that the two-timescale message passing mechanism effectively exploits live observations to enhance the fidelity of the impairment surrogate, which drives the channel estimation gain observed in Fig.~\ref{fig:online-varying-channel}.

\begin{figure}[!t]
\centering
\includegraphics[width=0.95\linewidth, clip]{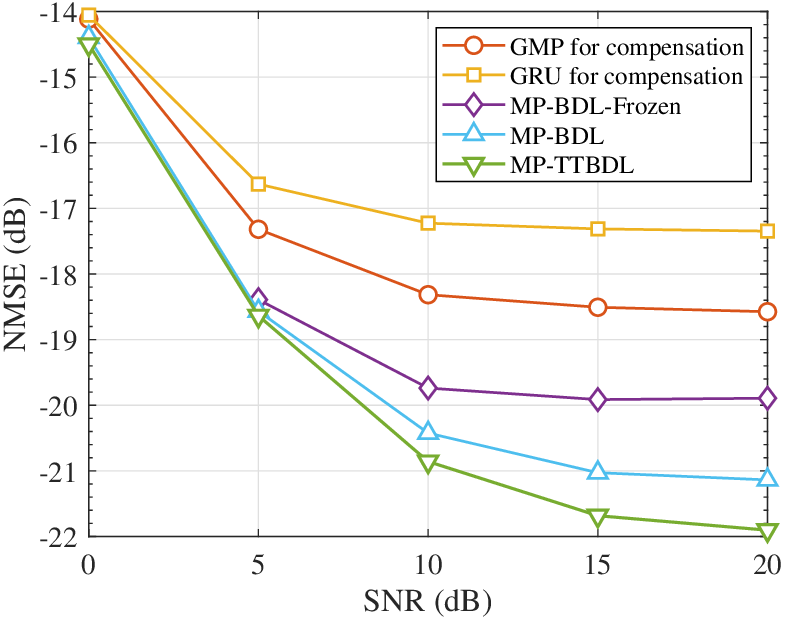}
\caption{Channel tracking NMSE versus SNR under time-varying impairments (averaged over the last $100$ slots).}
\label{fig:online-varying-channel-snr}
\end{figure}

Fig.~\ref{fig:online-varying-channel-snr} reports the channel estimation NMSE averaged over the last $100$ slots versus SNR. MP-TTBDL consistently achieves the lowest NMSE among all compared schemes, confirming its robustness. The performance gain is particularly pronounced at high SNR, since observations at higher SNR provide more informative messages for the online RGRU parameter update and thus amplify the advantage of the two-timescale tracking framework.

\section{Conclusion}
This paper proposed an MP-TTBDL framework for joint channel and hardware impairment tracking in massive MIMO receivers. An RGRU was adopted to capture the intra-slot memory of the impairments. Distinct Markov priors were placed on the fast-varying sparse channel and the slow-varying network parameters, allowing temporal messages to propagate across slots at two different timescales and thereby track both channel dynamics and hardware aging. To handle the complex intra-slot structure, the factor graph within each slot was split into a CT module and an IC module. Turbo-OAMP handled the sparse channel estimation, while a proposed DAMP procedure updated the impairment parameters, and the two modules refined their estimates by exchanging extrinsic information through EP. Specifically, a mean-field VBI scheme was employed to approximate the non-Gaussian messages produced by the nonlinear observation models. Simulation results demonstrated that the proposed framework achieves consistently lower channel estimation error than the baselines that apply compensation followed by Bayesian channel estimation across both static and slowly time-varying impairment conditions. This advantage is particularly pronounced at high SNR and accumulates over time during online tracking, while the per-slot computational complexity stays within a constant factor of these baselines.

Several directions deserve further investigation. First, extending the framework to multi-user scenarios, where multi-user pilot design may be jointly optimized to further improve the overall channel estimation performance, represents a natural generalization of this work. Second, more advanced recurrent architectures, e.g., Transformers, may be explored to address more complex hardware impairments expected in future transceivers. In such scenarios, message passing algorithms for the corresponding factor graphs should be developed. Third, since the per-slot computational cost of the proposed framework is dominated by the online tracking of slow-varying network parameters, low-complexity implementations are desirable for practical deployment to achieve a favorable tradeoff between tracking performance and algorithm complexity.

\appendices
\section{VBI Updates for Submodule $\boldsymbol{\pi}$}
\label{app:nonlinear-vbi}
This appendix derives the closed-form mean-field VBI updates for the nonlinear submodule \(\boldsymbol{\pi}\) in Sec.~\ref{subsec:nonlinear-vbi}.  The update formulas for \(\check{\boldsymbol{\pi}}\) are structurally similar and can be obtained by adapting the same procedure to its specific observation model (\ref{eq:picheck}).  Since the observation model is element-wise, we consider the derivation in the scalar case.  We denote by \(\hat{y}_{\mathrm{eq}}\) the equivalent observation with noise variance \(\sigma_{\mathrm{eq}}^2\).

\paragraph{Forward Updates (FMP)}
During FMP, the posterior distribution of $\pi_p$ is approximated by a Gaussian distribution. Since Eq.~(\ref{eq:pit}) is a deterministic mapping, the corresponding mean and variance are expressed based on the variational posteriors $q(z_p)$, $q(\tilde{\pi}_p)$, and $q(\pi_{p-1})$:
\begin{align*}
\mu_{\pi_p} &= \bigl(1-\mathbb{E}[\mathcal{Q}(z_p)]\bigr)\mu_{\pi_{p-1}} + \mathbb{E}[\mathcal{Q}(z_p)]\bigl(2\mathbb{E}[\mathcal{Q}(\tilde{\pi}_p)]-1\bigr), \\
v_{\pi_p} &= \bigl(1-\mathbb{E}[\mathcal{Q}(z_p)]\bigr)^2 v_{\pi_{p-1}} + \mu_{\pi_{p-1}}^2 \mathrm{Var}[\mathcal{Q}(z_p)] \\
&\quad + \mathrm{Var}[\mathcal{Q}(z_p)] v_{\pi_{p-1}} + \bigl(\mathbb{E}[\mathcal{Q}(z_p)]\bigr)^2 \cdot 4\mathrm{Var}[\mathcal{Q}(\tilde{\pi}_p)] \\
&\quad + \bigl(2\mathbb{E}[\mathcal{Q}(\tilde{\pi}_p)]-1\bigr)^2 \mathrm{Var}[\mathcal{Q}(z_p)] \\
&\quad + 4\mathrm{Var}[\mathcal{Q}(z_p)]\mathrm{Var}[\mathcal{Q}(\tilde{\pi}_p)],
\end{align*}
where for any Gaussian variable \(X\), the expectation \(\mathbb{E}[\mathcal{Q}(X)]\) and variance \(\mathrm{Var}[\mathcal{Q}(X)]\) can be expressed in terms of the bivariate normal cumulative distribution function, and the specific expressions can be found in~\cite{rasmussenGPML2006,genzBivariate2004}.
The forward message is \(\mathcal{N}(\mu_{\pi_p},v_{\pi_p})\), and is computed once after the BMP procedure below converges.

\paragraph{Backward Updates (BMP)}
Under the mean-field assumption, the variational distribution factorizes as \(q(z_p,\tilde{\pi}_p,\pi_{p-1}) = q(z_p)q(\tilde{\pi}_p)q(\pi_{p-1})\), with each factor constrained to a Gaussian form: \(q(z_p) = \mathcal{N}(z_p;\mu_{z_p},v_{z_p})\), \(q(\tilde{\pi}_p) = \mathcal{N}(\tilde{\pi}_p;\mu_{\tilde{\pi}_p},v_{\tilde{\pi}_p})\), and \(q(\pi_{p-1}) = \mathcal{N}(\pi_{p-1};\mu_{\pi_{p-1}},v_{\pi_{p-1}})\), where \(\mu_{z_p}\), \(v_{z_p}\), \(\mu_{\tilde{\pi}_p}\), \(v_{\tilde{\pi}_p}\), \(\mu_{\pi_{p-1}}\), and \(v_{\pi_{p-1}}\) denote the updated posterior parameters.  The equivalent priors are written as \(\mathcal{N}(\cdot;\mu_{z_p \rightarrow g_{\pi}^{t}},v_{z_p \rightarrow g_{\pi}^{t}})\), \(\mathcal{N}(\cdot;\mu_{\tilde{\pi}_p \rightarrow g_{\pi}^{t}},v_{\tilde{\pi}_p \rightarrow g_{\pi}^{t}})\), and \(\mathcal{N}(\cdot;\mu_{\pi_{p-1} \rightarrow g_{\pi}^{t}},v_{\pi_{p-1} \rightarrow g_{\pi}^{t}})\), respectively.  During BMP, the KL divergence between this variational distribution and the true posterior under (\ref{eq:pit}) is minimized with respect to the variational parameters.  For a given variable, e.g., \(\pi_{p-1}\), the optimal factor satisfies
\[
\log q(\pi_{p-1}) = \mathbb{E}_{q(z_p)q(\tilde{\pi}_p)}\bigl[\log p(\hat{y}_{\mathrm{eq}},z_p,\tilde{\pi}_p,\pi_{p-1})\bigr] + \text{const},
\]
where
\begin{flalign}
& p(\hat{y}_{\mathrm{eq}},z_p,\tilde{\pi}_p,\pi_{p-1}) && \nonumber \\
& = \mathcal{N}\bigl(\hat{y}_{\mathrm{eq}};\,(1-\mathcal{Q}(z_p))\pi_{p-1} + \mathcal{Q}(z_p)(2\mathcal{Q}(\tilde{\pi}_p)-1),\sigma_{\mathrm{eq}}^2\bigr) && \nonumber \\
& \times \Delta_{z_p \rightarrow g_{\pi}^{t}}\,\Delta_{\tilde{\pi}_p \rightarrow g_{\pi}^{t}}\,\Delta_{\pi_{p-1} \rightarrow g_{\pi}^{t}}, && \label{eq:joint}
\end{flalign}
Based on this expression, the update of \(\pi_{p-1}\) yields a Gaussian distribution whose mean and variance are obtained in closed form, whereas those of \(z_p\) and \(\tilde{\pi}_p\) require linearization of \(\mathcal{Q}(\cdot)\) around the current variational means.  The three factors are updated alternately until convergence, and the specific updating rules are provided below.
\begin{itemize}
\item \textbf{Update for \(\pi_{p-1}\):}
\begin{align*}
v_{\pi_{p-1}} &= \left(\frac{1}{v_{\pi_{p-1} \rightarrow g_{\pi}^{t}}}+\frac{1-2\mathbb{E}[\mathcal{Q}(z_p)]+\mathbb{E}[\mathcal{Q}(z_p)^2]}{\sigma_{\mathrm{eq}}^2}\right)^{-1}, \\
\mu_{\pi_{p-1}} &= v_{\pi_{p-1}}\Biggl(\frac{\mu_{\pi_{p-1} \rightarrow g_{\pi}^{t}}}{v_{\pi_{p-1} \rightarrow g_{\pi}^{t}}}+\frac{1}{\sigma_{\mathrm{eq}}^2}\bigl(\hat{y}_{\mathrm{eq}}\bigl(1-\mathbb{E}[\mathcal{Q}(z_p)]\bigr) \\
&\quad - \bigl(\mathbb{E}[\mathcal{Q}(z_p)]-\mathbb{E}[\mathcal{Q}(z_p)^2]\bigr)\bigl(2\mathbb{E}[\mathcal{Q}(\tilde{\pi}_p)]-1\bigr)\bigr)\Biggr).
\end{align*}

\item \textbf{Update for \(z_p\):}
Since the variational mean \(\mu_{z_p}\) varies slowly across successive iterations, a first-order Taylor expansion around \(\mu_{z_p}\) is sufficiently accurate, i.e.,
\[
\mathcal{Q}(z_p) \approx \mathcal{Q}(\mu_{z_p}) + \phi(\mu_{z_p})(z_p - \mu_{z_p}),
\]
where \(\phi(\cdot)\) denotes the standard normal probability density function.  With this approximation, the posterior mean and variance of \(z_p\) are given by
\begin{flalign*}
\quad & v_{z_p} = \Bigl(\frac{1}{v_{z_p\!\rightarrow\! g_{\pi}^{t}}}+\frac{\phi(\mu_{z_p})^2}{\sigma_{\mathrm{eq}}^2}\,\mathbb{E}\bigl[(2\mathcal{Q}(\tilde{\pi}_p)-1-\pi_{p-1})^2\bigr]\Bigr)^{-1} &&\\
\quad & \mu_{z_p} = v_{z_p}\,\frac{\mu_{z_p\!\rightarrow\! g_{\pi}^{t}}}{v_{z_p\!\rightarrow\! g_{\pi}^{t}}} + v_{z_p}\,\frac{\phi(\mu_{z_p})}{\sigma_{\mathrm{eq}}^2}\bigl(\mathcal{Q}(\mu_{z_p})-\phi(\mu_{z_p})\mu_{z_p}\bigr) && \\
\quad & \times \mathbb{E}\bigl[(2\mathcal{Q}(\tilde{\pi}_p)-1-\pi_{p-1})^2\bigr] && \\
\quad & + v_{z_p}\,\frac{\phi(\mu_{z_p})}{\sigma_{\mathrm{eq}}^2}\bigl(\hat{y}_{\mathrm{eq}}-\mathbb{E}[\pi_{p-1}]\bigr)\times \bigl(2\mathbb{E}[\mathcal{Q}(\tilde{\pi}_p)]-1\bigr) &&
\end{flalign*}

\item \textbf{Update for \(\tilde{\pi}_p\):}
Similarly, linearizing \(2\mathcal{Q}(\tilde{\pi}_p)-1\) around \(\mu_{\tilde{\pi}}\) gives
\begin{align*}
v_{\tilde{\pi}_p} &= \left(\frac{1}{v_{\tilde{\pi}_p \rightarrow g_{\pi}^{t}}}+\frac{\mathbb{E}[\mathcal{Q}(z_p)^2]\,(2\phi(\mu_{\tilde{\pi}_p}))^2}{\sigma_{\mathrm{eq}}^2}\right)^{-1}, \\
\mu_{\tilde{\pi}_p} &= v_{\tilde{\pi}_p}\Biggl(\frac{\mu_{\tilde{\pi}_p \rightarrow g_{\pi}^{t}}}{v_{\tilde{\pi}_p \rightarrow g_{\pi}^{t}}}+\frac{2\phi(\mu_{\tilde{\pi}_p})}{\sigma_{\mathrm{eq}}^2}\Bigl(\hat{y}_{\mathrm{eq}}-\mathbb{E}[\mathcal{Q}(z_p)]\mathbb{E}[\pi_{p-1}] \\
&\quad - \bigl(2\mathcal{Q}(\mu_{\tilde{\pi}_p})-1\bigr)\mathbb{E}[\mathcal{Q}(z_p)]\Bigr)\Biggr).
\end{align*}
\end{itemize}

The three updates are performed alternately until convergence. In our simulations, this process typically requires only two iterations.

\IEEEtriggeratref{21}
\bibliographystyle{IEEEtran}
\bibliography{DNN_CE_new}

\end{document}